\documentclass[11pt,a4paper]{article}
\usepackage{times,latexsym}
\usepackage{url}
\usepackage[T1]{fontenc}
\usepackage[hyperref]{acl2021}
\usepackage{xspace,mfirstuc,tabulary}
\usepackage{booktabs}
\usepackage{amssymb}
\usepackage{amsmath}
\usepackage{multirow,booktabs, hhline}
\usepackage[ruled,noend]{algorithm2e}
\usepackage{amsmath, bm}
\SetKwInput{KwInput}{Input}
\SetKwInput{KwOutput}{Output}
\usepackage{url}
\usepackage{graphicx}

\usepackage{color}                  % Text and background color

\usepackage{microtype}
\usepackage{scalerel}

\usepackage{tabularx}               % Better table formatting
\newcolumntype{C}{>{\centering\arraybackslash}X}
\usepackage{diagbox}                % Diagonal line in tables
\usepackage{mathtools}              % Formula
\usepackage{enumitem}               % Better itemize environment

\usepackage{subfigure}
\usepackage{booktabs}
\usepackage{extarrows}
\usepackage{makecell}

\usepackage{microtype}

\aclfinalcopy % Uncomment this line for the final submission
\usepackage{xparse}

\definecolor{OrangeRed}{rgb}{1.0, 0.27, 0.0}
\NewDocumentCommand{\heng}{ mO{} }{\textcolor{OrangeRed}{\textsuperscript{\textit{Heng}}\textsf{\textbf{\small[#1]}}}}

\newcommand\ourmodel{ERICA\xspace}

\title{\ourmodel: Improving Entity and Relation Understanding for \\ Pre-trained Language Models via Contrastive Learning}

\author{
 Yujia~Qin$^{\clubsuit\spadesuit\diamondsuit}$, Yankai~Lin$^{\diamondsuit}$, Ryuichi~Takanobu$^{\clubsuit\diamondsuit}$, Zhiyuan~Liu$^{\clubsuit*}$, Peng~Li$^\diamondsuit$, Heng~Ji$^{\spadesuit*}$, \\ 
 \textbf{Minlie Huang$^\clubsuit$, Maosong~Sun$^\clubsuit$, Jie~Zhou$^\diamondsuit$} \\
 $^\clubsuit$Department of Computer Science and Technology, Tsinghua University, Beijing, China \\
 $^\spadesuit$University of Illinois at Urbana-Champaign \\
 $^\diamondsuit$Pattern Recognition Center, WeChat AI, Tencent Inc. \\
 
\texttt{yujiaqin16@gmail.com}\\
}

\date{}
\begin{document}
\maketitle

\renewcommand{\thefootnote}{\fnsymbol{footnote}}
\footnotetext[1]{Corresponding author.}
\renewcommand{\thefootnote}{\arabic{footnote}}

\begin{abstract}

%\heng{since your experiments are beyond IE, consider to make the title broader}

Pre-trained Language Models (PLMs) have shown superior performance on various downstream Natural Language Processing (NLP) tasks. However, conventional pre-training objectives do not explicitly model relational facts in text, which are crucial for textual understanding. To address this issue, we propose a novel contrastive learning framework \ourmodel to obtain a deep understanding of the entities and their relations in text. Specifically, we define two novel pre-training tasks to better understand entities and relations: (1) the entity discrimination task to distinguish which tail entity can be inferred by the given head entity and relation; (2) the relation discrimination task to distinguish whether two relations are close or not semantically, which involves complex relational reasoning. Experimental results demonstrate that \ourmodel can improve typical PLMs (BERT and RoBERTa) on several language understanding tasks, including relation extraction, entity typing and question answering, especially under low-resource settings.\footnote{Our code and data are publicly available at \url{https://github.com/thunlp/ERICA}.}
% All datasets, source codes and model parameters will be available to advance further research explorations.
%We will release the datasets, source codes and pre-trained language models for further research explorations.
%Specially, \ourmodel  introduces both entity prediction and relation discrimination pre-training tasks to enhance %\textbf{Doc}umen\textbf{T}-level understanding for \textbf{R}elat\textbf{I}o\textbf{N}s and \textbf{E}ntities (\textbf{DocTRINE}). 
%there still exists challenges when extending these tasks to document level: (1) firstly, there may exist multiple entities in a document and relations between these entities can be very complex; (2) secondly, entities may occur multiple times throughout a document; (3) thirdly, the occurrences of entities do not contribute equally when considering the relations between two entities. Previous work has explored improving entity and relation representation via external knowledge bases, however, as is shown in our experiments, these methods can not well model interactions among multiple entities in documents. In this paper, we propose a general framework to enhance \textbf{Doc}umen\textbf{T}-level understanding for \textbf{R}elat\textbf{I}o\textbf{N}s and \textbf{E}ntities (\textbf{DocTRINE}) via contrastive learning.
\end{abstract}
\section{Introduction}
Pre-trained Language Models (PLMs)~\citep{devlin2018bert,yang2019xlnet,liu2019roberta} %such as BERT~\citep{devlin2018bert}, XLNet~\citep{yang2019xlnet} and RoBERTa~\citep{liu2019roberta}  
have shown superior performance on various Natural Language Processing (NLP) tasks such as text classification~\citep{wang2018glue}, named entity recognition~\citep{sang2003introduction}, and question answering~\citep{talmor2019multiqa}. %, among which, entity-centric tasks, such as entity typing and relation extraction, focus on detecting relevant information about entities in texts. 
%PLMs have served as cornerstones, especially for sentence-level entity-centric tasks where entity representations encoded by PLMs contain rich type information, relation information and other contextual information.
Benefiting from designing various effective self-supervised learning objectives, such as masked language modeling~\citep{devlin2018bert}, PLMs can effectively capture the syntax and semantics in text to generate informative language representations for downstream NLP tasks.

\begin{figure}[t!]
\centering
\includegraphics[width=0.45\textwidth]{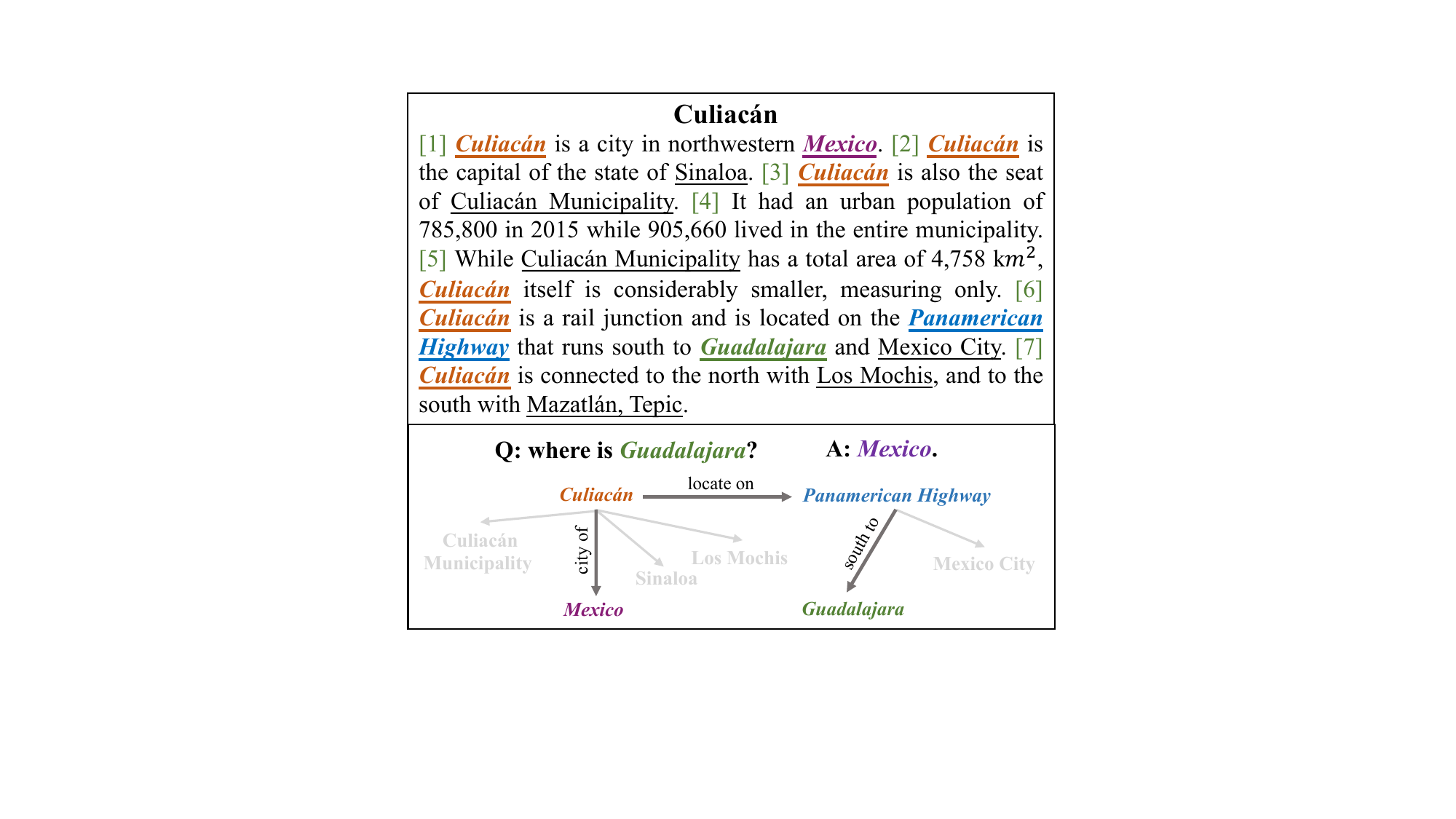}
\caption{An example for a document ``Culiacán'', in which all entities are underlined. We show entities and their relations as a relational graph, and highlight the important entities and relations to find out ``where is Guadalajara''. 
%\heng{so a tricky part about this example is that you also need to use commonsense knowledge that if two cities are near to each other as mentioned in the text, usually that indicate they belong to the same country.}
}
\label{fig:intra}
\end{figure}
% entities~\citep{joshi2019spanbert,xiong2019pretrained, zhang2019ernie, wang2019kepler, peters2019knowledge} or relations
However, conventional pre-training objectives do not explicitly model relational facts, which frequently distribute in text and are crucial for understanding the whole text. To address this issue, some recent studies attempt to improve PLMs to better understand relations between entities~\citep{soares2019matching, peng2020learning}. However, they mainly focus on within-sentence relations in isolation, ignoring the understanding of entities, and the interactions among multiple entities at document level, whose relation understanding involves complex reasoning patterns. According to the statistics on a human-annotated corpus sampled from Wikipedia documents by \citet{yao2019docred}, at least 40.7\% relational facts require to be extracted from multiple sentences. Specifically, we show an example in Figure~\ref{fig:intra}, to understand that ``Guadalajara is located in Mexico'', we need to consider the following clues jointly: (i) ``Mexico'' is the country of ``Culiacán'' from sentence 1; (ii) ``Culiacán'' is a rail junction located on ``Panamerican Highway'' from sentence 6; (iii)  ``Panamerican Highway'' connects to ``Guadalajara'' from sentence 6. From the example, we can see that there are two main challenges to capture the in-text relational facts: %\heng{consider to use different fonts for examples}
% 刘导：似乎与前面的interactions among multiple entities in documents没有太好照应。 从前面的论述看，还以为要处理多relational facts之间的复杂关联模式。需要调整前面说法。总之要将前后照应好。

1. To understand an entity, we should consider its relations to other entities comprehensively. In the example, the entity ``Culiacán'', occurring in sentence 1, 2, 3, 5, 6 and 7,  plays an important role in finding out the answer. To understand ``Culiacán'', we should consider all its connected entities and diverse relations among them.

% 刘导： Entity-Aware Knowledge Modeling. xxx
2. To understand a relation, we should consider the complex reasoning patterns in text. For example, to understand the complex inference chain in the example, we need to perform multi-hop reasoning, i.e., inferring that ``Panamerican Highway'' is located in ``Mexico'' through the first two clues. 
%and coreferential reasoning (inferring that the word ``It'' in sentence 7 refers to ``Panamerican Highway'').%\heng{did you do within-doc coreference in your framework?}
% 刘导： Relation-aware Knowledge Understanding.\

In this paper, we propose \ourmodel, a novel framework to improve PLMs' capability of \textbf{E}ntity and \textbf{R}elat\textbf{I}on understanding via \textbf{C}ontr\textbf{A}stive learning, aiming to better capture in-text relational facts by considering the interactions among entities and relations comprehensively. Specifically, we define two novel pre-training tasks: (1) the entity discrimination task to distinguish which tail entity can be inferred by the given head entity and relation. It improves the understanding of each entity via considering its relations to other entities in text; (2) the relation discrimination task to distinguish whether two relations are close or not semantically. Through constructing entity pairs with document-level distant supervision, it takes complex relational reasoning chains into consideration in an implicit way and thus improves relation understanding. %\heng{this paragraph is still very vague, make it more concrete by walking through Figure 1.}

We conduct experiments on a suite of language understanding tasks, including relation extraction, entity typing and question answering. The experimental results show that \ourmodel improves the performance of typical PLMs (BERT and RoBERTa) and outperforms baselines, especially under low-resource settings, which demonstrates that \ourmodel effectively improves PLMs' entity and relation understanding and captures the in-text relational facts.
\section{Related Work}

\citet{dai2015semi} and \citet{howard-ruder-2018-universal} propose to pre-train universal language representations on unlabeled text, and perform task-specific fine-tuning.
With the advance of computing power, PLMs such as OpenAI GPT~\citep{radford2018improving}, BERT~\citep{devlin2018bert} and XLNet~\citep{yang2019xlnet} based on deep Transformer~\citep{vaswani2017attention} architecture demonstrate their superiority in various downstream NLP tasks.
%With the advance of computing power, PLMs based on Transformer~\citep{vaswani2017attention} architecture demonstrate their superiority in learning universal language representations, such as the generative model OpenAI GPT~\citep{radford2018improving}, the deep bidirectional model BERT~\citep{devlin2018bert} and the permutation language model XLNet~\citep{yang2019xlnet}. 
Since then, numerous PLM extensions have been proposed to further explore the impacts of various model architectures~\citep{song2019mass,raffel2020exploring}, larger model size~\citep{raffel2020exploring,lan2019albert,fedus2021switch}, more pre-training corpora~\citep{liu2019roberta}, etc., to obtain better general language understanding ability. Although achieving great success, these PLMs usually regard words as basic units in textual understanding, ignoring the informative entities and their relations, which are crucial for understanding the whole text.

%Early studies of PLMs focus on learning distributed and static word representations~\citep{mikolov2013distributed, pennington2014glove} but struggle in dealing with polysemy. To this end, later work proposes to learn contextualized word embeddings~\citep{peters2018deep}. In spite of providing powerful word representations, these methods still require down-stream models to be learned from scratch. Another trend explores pre-training on unlabeled text and perform task-specific fine-tuning~\citep{dai2015semi,howard2018universal}. Recently, PLMs based on Transformer~\citep{vaswani2017attention} architecture demonstrate their superiority in learning universal language representations, such as generative models~\citep{radford2018improving}, deep bidirectional models~\citep{devlin2018bert} and permutation language models~\citep{yang2019xlnet}. Later work explores the impacts of more sophisticated pre-training~\citep{liu2019roberta} and more parameters~\citep{raffel2019exploring,lan2019albert}. However, these PLMs only focus on learning word tokens, ignoring informative entities and their relations, which are crucial for understanding the whole text.

To improve the entity and relation understanding of PLMs, a typical line of work is knowledge-guided PLM, which incorporates external knowledge such as Knowledge Graphs (KGs) into PLMs to enhance the entity and relation understanding. Some enforce PLMs to memorize information about real-world entities and propose novel pre-training objectives~\citep{xiong2019pretrained,wang2019kepler,sun-etal-2020-colake,yamada2020luke}. 
%such as zero-shot fact completion~\citep{xiong2019pretrained}, masked entity prediction~\citep{yamada2020luke} and knowledge embedding~\citep{wang2019kepler,sun-etal-2020-colake}.
Others modify the internal structures of PLMs to fuse both textual and KG's information~\citep{zhang2019ernie,peters2019knowledge,wang2020k,he-etal-2020-bert}.
%\citet{zhang2019ernie} add KG embeddings obtained by TransE~\citep{bordes2013translating} into PLMs to fuse both textual and KG's information. \citet{xiong2019pretrained} employ a zero-shot fact completion task and explicitly force PLMs to memorize entity information. \citet{yamada2020luke} propose to predict masked entities in a large entity-annotated corpus. Designed specifically for enhancing entity representations, these methods fail to consider the relational information between entities. \citet{peters2019knowledge} propose a general method to embed multiple KGs into PLMs and enhance their representations with structured, human-curated knowledge. Moreover, \citet{wang2019kepler} propose to encode entity descriptions with PLMs as entity embedding and learn them in the same way as conventional Knowledge Embedding methods.
Although knowledge-guided PLMs introduce extra factual knowledge in KGs, these methods ignore the intrinsic relational facts in text, making it hard to understand out-of-KG entities or knowledge in downstream tasks, let alone the errors and incompleteness of KGs. This verifies the necessity of teaching PLMs to understand relational facts from contexts.

%enhance PLMs' representations of continual spans, such as entity mentions, but ignore the interactions among different entities. \citet{soares2019matching} and \citet{peng2020learning} encourage relation representations of the same entity pair in different texts to be similar.

%\citet{peng2020learning} extend the above idea by grouping entity pairs that share the same relation annotated distantly by KGs. However, firstly, these methods focus on the task of relation extraction, ignoring the understanding of entities; secondly, they only explore capturing within-sentence relations, which limits the performance in dealing with multiple entities and relations at document level.
%\heng{if these are two disadvantages, don't use 'on one hand, on the other hand' which usually means two contrastive facts.}

Another line of work is to directly model entities or relations in text in pre-training stage to break the limitations of individual token representations. Some focus on obtaining better span representations, including entity mentions, via span-based pre-training~\citep{sun2019ernie,joshi2019spanbert,kong2019mutual,ye2020coreferential}. Others learn to extract relation-aware semantics from text by comparing the sentences that share the same entity pair or distantly supervised relation in KGs~\citep{soares2019matching,peng2020learning}. However, these methods only consider either individual entities or within-sentence relations, which limits the performance in dealing with multiple entities and relations at document level. In contrast, our \ourmodel considers the interactions among multiple entities and relations comprehensively, achieving a better understanding of in-text relational facts.

%However, although these methods introduce extra factual knowledge in KGs, restricted by the over-reliance on KG annotation, the PLMs themselves are not trained to fully capture the relational facts within the context. This situation is more severe when dealing with out-of-KG entities in downstream fine-tuning.

%In this work, we aim to improve PLMs' understanding of entities and relations to better capture in-text relational facts for language understanding.

%performance on document-level tasks unsatisfying. 
%\citet{xiao2020denoising} explores denoising relation extraction from document-level distant supervision. 
%However, their method requires that an annotated document-level RE dataset is provided, making it hard to be generalized in low resource setting.

\section{Methodology}

In this section, we introduce the details of \ourmodel. We first describe the notations and how to represent entities and relations in documents. Then we detail the two novel pre-training tasks: Entity Discrimination (ED) task and Relation Discrimination (RD) task, followed by the overall training objective.

%1. \textbf{Entity Discrimination} (ED) task that improves PLMs' entity representations via considering the diverse relations among them.

%2. \textbf{Relation Discrimination} (RD) task that improves PLMs' relation representations via additionally considering the complex reasoning chains.

%Relation Prediction (RP) task and . Both tasks train PLMs to understand entities and relations via tackling the challenges of information aggregation and propagation from different angles.  

%entity and relation representations via document-level contrastive pre-training. Contrastive learning~\citep{hadsell2006dimensionality} learns representations by grouping ``neighbors'' together and pushing ``non-neighbors'' apart. We solicit to distant supervision \citep{mintz2009distant} for defining ``neighbors'' in each task. As stated in the previous section, \ourmodel does not change the inner structures of PLMs nor introduce additional neural structures, making it convenient to be applied on various downstream tasks. We first describe how to represent entities and relations in documents and then detail the two contrastive learning-based tasks: Relation Prediction (RP) task and Entity Prediction (EP) task. Both tasks train PLMs to understand entities and relations via tackling the challenges of information aggregation and propagation from different angles.

\begin{figure}[!t]
\centering
\includegraphics[width=0.45\textwidth]{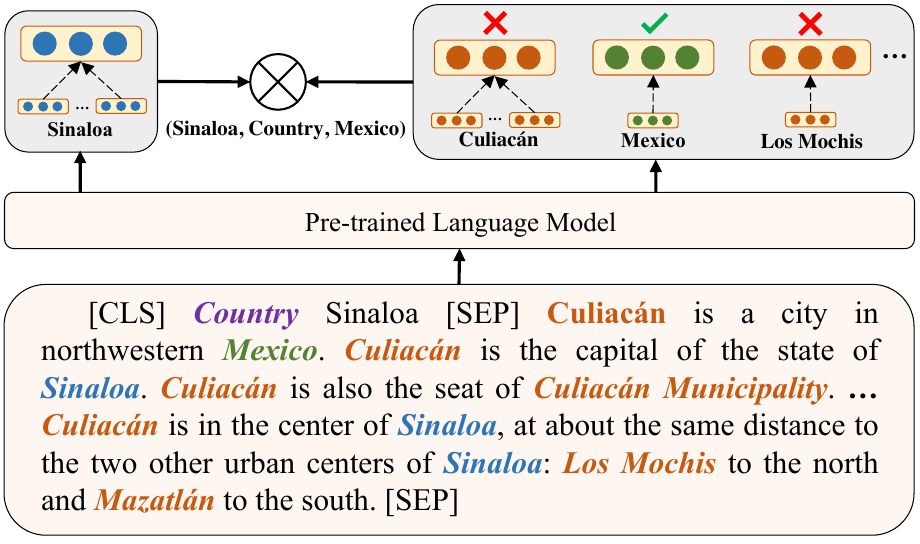}
\caption{An example of Entity Discrimination task. For an entity pair with its distantly supervised relation in text, %we concatenate the relation name and the head entity mention in front of the text as a hint for PLMs to know what to be conditioned on. 
the ED task requires the ground-truth tail entity to be closer to the head entity than other entities. %Selected instances contain both single-sentence and cross-sentence entity pairs.
}
\label{fig:EP_task}
\end{figure}

\subsection{Notations}
\ourmodel is trained on a large-scale unlabeled corpus leveraging the distant supervision from an external KG $\mathcal{K}$. Formally, let $\mathcal{D} = \{d_i\}_{i=1}^{|\mathcal{D}|}$ be a batch of documents and $\mathcal{E}_i =\{e_{ij}\}_{j=1}^{|\mathcal{E}_i|}$ be all named entities in $d_i$, where $e_{ij}$ is the $j$-th entity in $d_i$. For %Given an external knowledge base $\mathcal{K}$ containing a relation set $\mathcal{R}$, we first recognize all named entities $\mathcal{E}_i$ within 
each document $d_i$, we enumerate all entity pairs $(e_{ij}, e_{ik})$ and link them to their corresponding relation $r^i_{jk}$ in $\mathcal{K}$ (if possible) and obtain a tuple set $\mathcal{T}_i $ = $ \{t_{jk}^{i} = (d_i, e_{ij}, r^i_{jk}, e_{ik}) | j \neq k\}$. We assign \texttt{no\_relation} to those entity pairs without relation annotation in $\mathcal{K}$. Then we obtain the overall tuple set $\mathcal{T} = \mathcal{T}_1\bigcup\mathcal{T}_2 \bigcup ...\bigcup \mathcal{T_{|\mathcal{D}|}}$ for this batch. The positive tuple set $\mathcal{T}^{+}$ is constructed by removing all tuples with \texttt{no\_relation} from $\mathcal{T}$. Benefiting from document-level distant supervision, $\mathcal{T}^{+}$ includes both intra-sentence (relatively simple cases) and inter-sentence entity pairs (hard cases), whose relation understanding involves cross-sentence, multi-hop, or coreferential reasoning, i.e., $\mathcal{T}^{+} = \mathcal{T}^{+}_{single}\bigcup\mathcal{T}^{+}_{cross}$.

%   where $r^i_{jk}$ is the corr relation between $e_{ij}$ and $e_{ik}$. Note $(e_{ij}, e_{ik})$ and $(e_{ik}, e_{ij})$ may have different relations. After querying $\mathcal{K}$, each entity pair is either linked with a pre-defined relation $r^i_{jk} \in \mathcal{R}$ if possible, %i.e., $(e_{ij}, r^i_{jk}, e_{ik}) \in \mathcal{K}$, 
%or \texttt{no\_relation} otherwise. %, i.e., $r^i_{jk} =$ \texttt{no\_relation}. 

\subsection{Entity \& Relation Representation}
\label{ER_rep}
% Let $\mathcal{D} = \{d_i\}_{i=1}^{|\mathcal{D}|}$ be $|\mathcal{D}|$ documents, with each document $d_i = (\omega_{i1}, \omega_{i2}, ..., \omega_{in})$ containing $n$ tokens. Let $\mathcal{E}_i =\{e_{ij}\}_{j=1}^{|\mathcal{E}_i|}$ be all named entities in $d_i$, where $e_{ij}$ is the j-th entity in $d_i$. Since $e_{ij}$ may appear multiple times in $d_i$, possibly with slightly different mentions, we denote the set of all occurrences of $e_{ij}$ as $\mathcal{S}_{e_{ij}} = \{m^k_{e_{ij}}\}_{k=1}^{|S_{e_{ij}}|}$, where each occurrence $m^k_{e_{ij}}$ contains several tokens in $d_i$ from index $n^k_1$ to $n^k_2$ ($1 \leq n^k_1 \leq n^k_2 \leq n$), i.e., $m^k_{e_{ij}}$ = ($\omega_{in^k_1}$, ..., $\omega_{in^k_2}$). We first use PLM to encode $d_i$ and get a series of hidden states $(\mathbf{h}_1, ..., \mathbf{h}_n)$, then we apply mean pooling over all token representations $(\mathbf{h}_{n^k_1}, ..., \mathbf{h}_{n^k_2})$ to represent $m^k_{e_{ij}}$ as $\mathbf{m}^k_{e_{ij}}$, which denotes the local contextual information of $e_{ij}$ at k-th occurrence:

For each document $d_i$, we first use a PLM to encode it and obtain a series of hidden states $\{\mathbf{h}_1, \mathbf{h}_2, ..., \mathbf{h}_{|d_i|}\}$, then we apply \textit{mean pooling} operation over the consecutive tokens that mention $e_{ij}$ to obtain local entity representations. Note $e_{ij}$ may appear multiple times in $d_i$, the $k$-th occurrence of $e_{ij}$, which contains the tokens from index $n^k_{start}$ to $n^k_{end}$, is represented as:
\begin{equation}
% \small
\begin{aligned}
\mathbf{m}^k_{e_{ij}} &= \text{MeanPool} (\mathbf{h}_{n^k_{start}}, ..., \mathbf{h}_{n^k_{end}}). \\
\end{aligned}
\label{eq:represent}
\end{equation}

To aggregate all information about $e_{ij}$, we average\footnote{Although weighted summation by attention mechanism is an alternative, the specific method of entity information aggregation is not our main concern.} all representations of each occurrence $\mathbf{m}^k_{e_{ij}}$ as the global entity representation $\mathbf{e}_{ij}$.
% \begin{equation}
% % \small
% \begin{aligned}
% \mathbf{e}_{ij} &= \frac{1}{K_{e_{ij}}}\sum_{k=1}^{K_{e_{ij}}} \mathbf{m}^k_{e_{ij}}. \\
% \end{aligned}
% \end{equation}
% where we denote the number of all occurrences of $e_{ij}$ as $K_{e_{ij}}$.
Following~\citet{soares2019matching}, we concatenate the final representations of two entities $e_{ij_1}$ and $e_{ij_2}$ as their relation representation, i.e., $\mathbf{r}^i_{j_1j_2} = [\mathbf{e}_{ij_1}; \mathbf{e}_{ij_2}]$.% We construct relation sets $\mathcal{R}$ and $\mathcal{R}^{+}$, which will be used in RD task, by retaining the relation representation of each tuple in $\mathcal{T}$ and $\mathcal{T}^{+}$, respectively.
%$\mathbf{r}^i_{j_1j_2} = \langle \mathbf{e}_{ij_1}|\mathbf{e}_{ij_2} \rangle$, where $\langle \cdot | \cdot \rangle$ denotes concatenation.

\subsection{Entity Discrimination Task}
\label{EP_task}

Entity Discrimination (ED) task aims at inferring the tail entity in a document given a head entity and a relation. By distinguishing the ground-truth tail entity from other entities in the text, it teaches PLMs to understand an entity via considering its relations with other entities. %EP task is fully defined on one document and thus does not involve inter-document interactions. 

%Next we obtain $\mathcal{T'}_i$ by excluding the triples in $\mathcal{T}_i$ that $r^i_{jk} = \texttt{no\_relation}$. Then we merge all the triple sets and get $\mathcal{T} = \{\mathcal{T}_1, ..., \mathcal{T_{|\mathcal{D}|}}\}$ and $\mathcal{T'} = \{\mathcal{T'}_1, ..., \mathcal{T'}_{|\mathcal{D}|}\}$, respectively. 
As shown in Figure~\ref{fig:EP_task}, in practice, we first sample a tuple $t_{jk}^{i}$ = $(d_i, e_{ij}$, $r^i_{jk}$, $e_{ik})$ from $\mathcal{T}^+$, PLMs are then asked to distinguish the ground-truth tail entity $e_{ik}$ from other entities in the document $d_i$.
%find $e_{ik}$ given $e_{ij}$ and $r^i_{jk}$, 
%which is equivalent to maximizing the posterior $\mathcal{P}(e_{ik} |e_{ij}, r^i_{jk})$. 
To inform PLMs of which head entity and relation to be conditioned on, we concatenate the relation name of $r_{jk}^i$, the mention of head entity $e_{ij}$ and a separation token \texttt{[SEP]} in front of $d_i$, i.e., $d_i^* = $``\texttt{relation\_name} \texttt{entity\_mention}\texttt{[SEP]} $d_i$''\footnote{Here we encode the modified document $d_i^*$ to obtain the entity representations. The newly added $\texttt{entity\_mention}$ is not considered for head entity representation.}.
%Under this situation, we call $e_{ij}$ and $e_{ik}$ entity-level ``neighbors'', while other entities in $d_i$ are ``non-neighbors'' of $e_{ik}$. Intuitively, the entity representations between ``neighbors'' should be closer than those of ``non-neighbors''.
The goal of entity discrimination task is equivalent to maximizing the posterior $\mathcal{P}(e_{ik} |e_{ij}, r^i_{jk}) = \text{softmax}(f(\mathbf{e}^{}_{ik}))$ ($f(\cdot)$ indicates an entity classifier). However, we empirically find directly optimizing the posterior cannot well consider the relations among entities. Hence, we borrow the idea of contrastive learning~\citep{hadsell2006dimensionality} and push the representations of positive pair ($e_{ij}$, $e_{ik}$) closer than negative pairs, the loss function of ED task can be formulated as:
% \begin{equation}
% \begin{aligned}
% \label{EP_loss}
%   s^e_{jk} &= \exp(\cos(\mathbf{e}^{}_{ij}, \mathbf{e}^{}_{ik}) / \tau), \\
%   \mathcal{L}_{\text{ED}} &= - \log \frac{s^e_{jk}}{s^e_{jk} + \sum\limits_{1\leq l \leq |\mathcal{E}_i|,\; l\notin \{j, k\}} s^e_{jl}},
% \end{aligned}
% \end{equation}
\begin{equation}
\begin{aligned}
\label{EP_loss}
  \mathcal{L}_{\text{ED}} = - \sum_{t_{jk}^{i}\in \mathcal{T}^+} \log \frac{\exp(\cos(\mathbf{e}^{}_{ij}, \mathbf{e}^{}_{ik}) / \tau)}{\sum\limits_{l=1,\; l\neq j}^{|\mathcal{E}_i|} \exp(\cos(\mathbf{e}^{}_{ij}, \mathbf{e}^{}_{il}) / \tau)},
  %,\; l\neq j
\end{aligned}
\end{equation}
where $\cos(\cdot, \cdot)$ denotes the cosine similarity between two entity representations and $\tau$ (temperature) is a hyper-parameter. %All recognized entities in $d_i^*$ except for $e_{ij}$ and $e_{ik}$, if not having the same relation $r_{jk}^i$ with $e_{ij}$, serve as negative samples.

\subsection{Relation Discrimination Task}
\label{RP_task}
\begin{figure}[!t]
\centering
\includegraphics[width=0.45\textwidth]{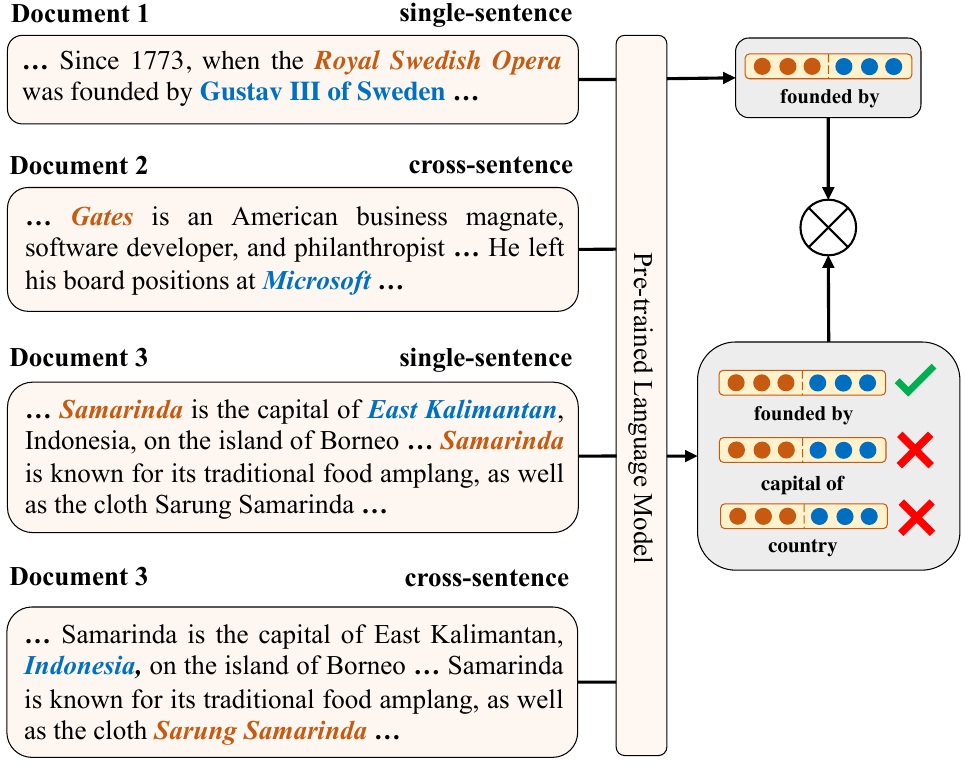}
\caption{An example of Relation Discrimination task. For entity pairs belonging to the same relations, the RD task requires their relation representations to be closer. %\heng{make font bigger in the figure. This example does not seem challenging enough.}
}
%are sampled from documents and their relationships are labeled distantly according to external Knowledge Base. We use PLMs to encode each document and obtain the relation representations, then these relation representations are trained by a contrastive learning objective. }
%Selected instances contain both single-sentence and cross-sentence entity pairs.}
\label{fig:RP_task}
\end{figure}
Relation Discrimination (RD) task aims at distinguishing whether two relations are close or not semantically. Compared with existing relation-enhanced PLMs, we employ document-level rather than sentence-level distant supervision to further make PLMs comprehend the complex reasoning chains in real-world scenarios and thus improve PLMs' relation understanding.

%Compared with sentence-level instance selection, which requires that two entities must co-occur in a sentence and thus confines the scale of training data, document-level instance selection includes both intra-sentence (relatively simple cases) and inter-sentence relational pairs, which may involve cross-sentence, multi-hop, or coreferential reasoning. Thus it has better coverage and generality.

As depicted in Figure \ref{fig:RP_task}, we train the text-based relation representations of the entity pairs that share the same relations to be closer in the semantic space. In practice, we linearly\footnote{The sampling rate of each relation is proportional to its total number in the current batch.} sample a tuple pair $t_A$ = ($d_A, e_{A_1}$, $r_A$, $e_{A_2}$) and $t_B$ = ($d_B, e_{B_1}$, $r_B$, $e_{B_2}$) from $\mathcal{T}^{+}_{s}$ ($\mathcal{T}^{+}_{single}$) or $\mathcal{T}^{+}_{c}$ ($\mathcal{T}^{+}_{cross}$), where $r_A = r_B$. Using the method mentioned in Sec.~\ref{ER_rep}, we obtain the positive relation representations $\mathbf{r}_{t_A}$ and $\mathbf{r}_{t_B}$ for $t_A$ and $t_B$. To discriminate positive examples from negative ones, similarly, we adopt contrastive learning and define the loss function of RD task as follows:
% \begin{equation}
% \begin{aligned}
% \label{RP_loss}
%   s^r_{AB} &= \exp(\cos(\mathbf{r}_{A}, \mathbf{r}_{B})/\tau), \\
%   \mathcal{L}_{\text{RD}} &= - \log \frac{s^r_{AB}}{ s^r_{AB} + \sum\limits_{C\in \mathcal{T}/\{A, B\}} s^r_{AC}}.
% \end{aligned}
% \end{equation}
\begin{equation}
\begin{aligned}
\label{RP_loss}
%   \mathcal{L}_{\text{RD}} = - \log \frac{\exp(\cos(\mathbf{r}_{A}, \mathbf{r}_{B})/\tau)}{ \sum\limits_{C\in \mathcal{T}}^{N_{neg}} \exp(\cos(\mathbf{r}_{A}, \mathbf{r}_{C})/\tau)},
    \mathcal{L}_{\text{RD}}^{\scriptscriptstyle \mathcal{T}_{1}, \mathcal{T}_{2}} &= - \sum_{t_A\in \mathcal{T}_{1}, t_B\in \mathcal{T}_{2}} \log \frac{\exp(\cos(\mathbf{r}_{t_A}, \mathbf{r}_{t_B})/\tau)}{\mathcal{Z}} ,\\
    \mathcal{Z} &= \sum\limits_{t_C\in \mathcal{T}/\{t_A\}}^{N} \exp(\cos(\mathbf{r}_{t_A}, \mathbf{r}_{t_C})/\tau) ,\\
    \mathcal{L}_{\text{RD}} &=  \mathcal{L}_{\text{RD}}^{ \mathcal{T}^{+}_{s}, \mathcal{T}^{+}_{s}} + \mathcal{L}_{\text{RD}}^{\scriptscriptstyle \mathcal{T}^{+}_{s}, \mathcal{T}^{+}_{c}} + \mathcal{L}_{\text{RD}}^{\scriptscriptstyle \mathcal{T}^{+}_{c}, \mathcal{T}^{+}_{s}} +
    \mathcal{L}_{\text{RD}}^{\scriptscriptstyle \mathcal{T}^{+}_{c}, \mathcal{T}^{+}_{c}}, \\
    % \sum_{\mathcal{T}_A, \mathcal{T}_B\in \mathcal{T}^{+}_{single}} \log \frac{\exp(\cos(\mathbf{r}_{A}, \mathbf{r}_{B})/\tau)}{\mathcal{Z}}\\
    %  -& \sum_{\mathcal{T}_A\in \mathcal{T}^{+}_{single}, \mathcal{T}_B\in \mathcal{T}^{+}_{cross}} \log \frac{\exp(\cos(\mathbf{r}_{A}, \mathbf{r}_{B})/\tau)}{\mathcal{Z}}\\
    %  -& \sum_{\mathcal{T}_A, \mathcal{T}_B\in \mathcal{T}^{+}_{cross}} \log \frac{\exp(\cos(\mathbf{r}_{A}, \mathbf{r}_{B})/\tau)}{\mathcal{Z}}\\
    % \mathcal{L}_{\text{RD}} &= - \sum_{\mathcal{T}_A, \mathcal{T}_B\in \mathcal{T}^{+}_{single}} \log \frac{\exp(\cos(\mathbf{r}_{A}, \mathbf{r}_{B})/\tau)}{\mathcal{Z}}\\
    %  -& \sum_{\mathcal{T}_A\in \mathcal{T}^{+}_{single}, \mathcal{T}_B\in \mathcal{T}^{+}_{cross}} \log \frac{\exp(\cos(\mathbf{r}_{A}, \mathbf{r}_{B})/\tau)}{\mathcal{Z}}\\
    %  -& \sum_{\mathcal{T}_A, \mathcal{T}_B\in \mathcal{T}^{+}_{cross}} \log \frac{\exp(\cos(\mathbf{r}_{A}, \mathbf{r}_{B})/\tau)}{\mathcal{Z}}\\
    %  \mathcal{Z} &= \sum\limits_{C\in \mathcal{T}}^{N_{neg}} \exp(\cos(\mathbf{r}_{A}, \mathbf{r}_{C})/\tau)\\ %/\{A\}
\end{aligned}
\end{equation}
where $N$ is a hyper-parameter. We ensure $t_B$ is sampled in $\mathcal{Z}$ and construct $N - 1$ negative examples by sampling $t_C$ ($r_A \neq r_C$) from $\mathcal{T}$, instead of $\mathcal{T}^{+}$\footnote{In experiments, we find introducing \texttt{no\_relation} entity pairs as negative samples further improves the performance and the reason is that increasing the diversity of training entity pairs is beneficial to PLMs.}. By additionally considering the last three terms of $\mathcal{L}_{\text{RD}}$ in Eq.\ref{RP_loss}, which require the model to distinguish complex inter-sentence relations with other relations in the text, our model could have better coverage and generality of the reasoning chains. PLMs are trained to perform reasoning in an implicit way to understand those ``hard'' inter-sentence cases.

\subsection{Overall Objective}
% \begin{algorithm}[!t]
%         \small
%         \caption{Joint Training Algorithm of \ourmodel}
%     \KwInput{raw corpus and Knowledge Base $\mathcal{K}$.}
%     \KwOutput{pre-trained language model $\mathcal{F}$.}
%     \While{not converge}{
%         Sample $N_{RP}$ documents  $\mathcal{D}_{RP}$ and $N_{EP}$ documents $\mathcal{D}_{EP}$. \\
%         Get tuple sets $\mathcal{T}_{RP}, \mathcal{T}_{EP}$ and $\mathcal{T'}_{RP}, \mathcal{T'}_{EP}$ by querying $\mathcal{K}$ as is described in \ref{RP_task}. \\ 
%         \ForEach{$d_i \in \mathcal{D}_{EP}$}{
%         Sample a triple $(e_{ij}, r^i_{jk}, e_{ik})$ from $d_i$, concatenate the relation name of $r^i_{jk}$ and mention of $e_{ij}$ in front of $d_i$. \\
%         }
%         Calculate $\mathcal{L}_{RP}$ on $\mathcal{D}_{RP}$ using Eq. \ref{RP_loss}. \\ 
%         Calculate $\mathcal{L}_{EP}$ on the modified $\mathcal{D}_{EP}$ using Eq. \ref{EP_loss}. \\ 
%         Calculate $\mathcal{L}_{MLM}$ on $\mathcal{D}_{RP}$ and $\mathcal{D}_{EP}$. \\ 
%         Calculate $\mathcal{L}$ using Eq. \ref{joint} and update $\mathcal{F}$. \\
%     }
%     \label{alg}
% \end{algorithm}

Now we present the overall training objective of \ourmodel. To avoid catastrophic forgetting~\citep{mccloskey1989catastrophic} of general language understanding ability, we train masked language modeling task ($\mathcal{L}_{\text{MLM}}$) together with ED and RD tasks. Hence, the overall learning objective is formulated as follows:

\begin{equation}
\begin{aligned}
\label{joint}
  \mathcal{L} = \mathcal{L}_{\text{ED}} + \mathcal{L}_{\text{RD}} + \mathcal{L}_{\text{MLM}}.
\end{aligned}
\end{equation}

%As is described above, $e_i$ may appear multiple times in $d$, although we average all these occurrences, we also tried introducing additional neural networks to implement attention mechanism for weighted summation over all occurrences of $e_i$. However we find averaging all these occurrences is always better. We hypothesize that the complex multi-head attention already conduct attention mechanism over these occurrences.

It is worth mentioning that we also try to mask entities as suggested by \citet{soares2019matching} and \citet{peng2020learning}, aiming to avoid simply relearning an entity linking system. However, we do not observe performance gain by such a masking strategy. We conjecture that in our document-level setting, it is hard for PLMs to overfit on memorizing entity mentions due to the better coverage and generality of document-level distant supervision. Besides, masking entities creates a gap between pre-training and fine-tuning, which may be a shortcoming of previous relation-enhanced PLMs.
\section{Experiments}

%\heng{I really think you should put more qualitative analysis by providing examples.}

%\heng{This section is very dry. Add qualitative analysis with real output examples, add discussions on error analysis and remaining challenges. Let me know when you are done and then I'll review/comment on this section.}
In this section, we first describe how we construct the distantly supervised dataset and pre-training details for \ourmodel. Then we introduce the experiments we conduct on several language understanding tasks, including relation extraction (RE), entity typing (ET) and question answering (QA). 
%We encode each entity by using mean pooling over all occurrences of it. 
%For tasks that do not provide the index of each entity, we find entities by exact matching in texts.
We test \ourmodel on two typical PLMs, including BERT and RoBERTa (denoted as $\text{\ourmodel}_{\texttt{BERT}}$ and $\text{\ourmodel}_{\texttt{RoBERTa}}$)\footnote{Since our main focus is to demonstrate the superiority of \ourmodel in improving PLMs to capture relational facts and advance further research explorations, we choose base models for experiments.}. We leave the training details for downstream tasks and experiments on GLUE benchmark~\citep{wang2018glue} in the appendix.

\subsection{Distantly Supervised Dataset Construction}
  \label{distant_pretraining_data}
Following \citet{yao2019docred}, we construct our pre-training dataset leveraging distant supervision from the English Wikipedia and Wikidata. First, we use spaCy\footnote{\url{https://spacy.io/}} to perform \textit{Named Entity Recognition}, and then link these entity mentions as well as Wikipedia's mentions with hyper-links to Wikidata items, thus we obtain the Wikidata ID for each entity. The relations between different entities are annotated distantly by querying Wikidata. We keep the documents containing at least $128$ words, $4$ entities and $4$ relational triples. In addition, we ignore those entity pairs appearing in the test sets of RE and QA tasks to avoid test set leakage. In the end, we collect $1,000,000$ documents (about 1G storage) in total with more than $4,000$ relations annotated distantly. On average, each document contains $186.9$ tokens, $12.9$ entities and $7.2$ relational triples, an entity appears $1.3$ times per document. Based on the human evaluation on a random sample of the dataset, we find that it achieves an F1 score of 84.7\% for named entity recognition, and an F1 score of 25.4\% for relation extraction.

\subsection{Pre-training Details}
We initialize $\text{\ourmodel}_{\texttt{BERT}}$ and $\text{\ourmodel}_{\texttt{RoBERTa}}$ with \textit{bert-base-uncased} and \textit{roberta-base} checkpoints released by Google\footnote{\url{https://github.com/google-research/bert}} and Huggingface\footnote{\url{https://github.com/huggingface/transformers}}. We adopt AdamW~\citep{loshchilov2017decoupled} as the optimizer, warm up the learning rate for the first $20$\% steps and then linearly decay it. We set the learning rate to $3\times 10^{-5}$, weight decay to $1\times 10^{-5}$, batch size to $2,048$ and temperature $\tau$ to $5\times 10^{-2}$. For $\mathcal{L}_\text{RD}$, we randomly select up to $64$ negative samples per document. We train both models with $8$ NVIDIA Tesla P40 GPUs for $2,500$ steps.

%while for $L_{EP}$, all named entities identified in the document serve as negative samples

\subsection{Relation Extraction}

Relation extraction aims to extract the relation between two recognized entities from a pre-defined relation set. We conduct experiments on both document-level and sentence-level RE. We test three partitions of the training set ($1$\%, $10$\% and $100$\%) and report results on test sets.

\begin{table}[!t]
  \centering
  \small
    \begin{tabular}{l@{~~~}c@{~~~}c@{~~~}c@{~~~}c@{~~~}c@{~~~}c}
    \toprule
    \textbf{Size}      & \multicolumn{2}{c}{1\%}      & \multicolumn{2}{c}{10\%}      & \multicolumn{2}{c}{100\%} \\
    \midrule
    \textbf{Metrics} & F1    & IgF1  & F1    & IgF1  & F1    & IgF1 \\
    \midrule
    % \multicolumn{4}{l}{\textbf{Traditional Models}}\\
    CNN   & \multicolumn{2}{c}{-}         & \multicolumn{2}{c}{-}  & 42.3  & 40.3  \\
    BILSTM & \multicolumn{2}{c}{-}         & \multicolumn{2}{c}{-} & 51.1  & 50.3  \\
    \midrule
    % \multicolumn{4}{l}{\textbf{BERT-based Models}}\\
    BERT    & 30.4  & 28.9   & 47.1  & 44.9  & 56.8  & 54.5  \\
    HINBERT & \multicolumn{2}{c}{-}         & \multicolumn{2}{c}{-}        & 55.6  & 53.7  \\
    CorefBERT  & 32.8  & 31.2   & 46.0  & 43.7   & 57.0  & 54.5  \\
    SpanBERT   & 32.2  & 30.4   & 46.4  & 44.5   & 57.3  & 55.0  \\
    ERNIE   &  26.7  &  25.5      &  46.7  &  44.2    &  56.6   & 54.2 \\
    MTB    & 29.0  & 27.6   & 46.1  & 44.1   & 56.9  & 54.3  \\
    CP      & 30.3  & 28.7    & 44.8  & 42.6    & 55.2  & 52.7  \\
    %\midrule
    %$\text{\ourmodel}_{\texttt{BERT}}$ - $L_{EP}$ & 50.4  & 48.4  & 50.4  & 48.2  & 56.7  & 54.7  & 56.1  & 54.0  & 57.9  & 55.7  & 57.9  & 55.6  \\
    %$\text{\ourmodel}_{\texttt{BERT}}$ - $L_{RP}$ & 49.9  & 47.8  & 49.8  & 47.6  & 55.7  & 53.3  & 56.4  & 54.0  & 58.0  & 55.7  & 58.2  & 55.8  \\
    $\text{\ourmodel}_{\texttt{BERT}}$   & \textbf{37.8}  & \textbf{36.0}   & \textbf{50.8}  & \textbf{48.3}    & \textbf{58.2}  & \textbf{55.9}  \\
    \midrule
    % \multicolumn{4}{l}{\textbf{RoBERTa-based Models}}\\
    RoBERTa & 35.3  & 33.5   & 48.0  & 45.9    & 58.5  & 56.1  \\
    $\text{\ourmodel}_{\texttt{RoBERTa}}$   & \textbf{40.1}  & \textbf{38.0}   & \textbf{50.3}  & \textbf{48.3}   & \textbf{59.0}  & \textbf{56.6}  \\
    \bottomrule
    \end{tabular}%
  \caption{Results on document-level RE (DocRED). We report micro F1 (F1) and micro ignore F1 (IgF1) on test set. IgF1 metric ignores the relational facts shared by the train and dev/test sets.}
  \label{tab:docred}
\end{table}%

\begin{table}[!tbp]
  \centering
  \small
    \begin{tabular}{l@{~~~}c@{~~~}c@{~~~}c@{~~~}c@{~~~}c@{~~~}c@{~~~}c@{~~~}c}
        \toprule
    \textbf{Dataset}     & \multicolumn{3}{c}{TACRED} & \multicolumn{3}{c}{SemEval} \\
    \midrule
    \textbf{Size}  & 1\%   & 10\%  & 100\% & 1\%   & 10\%  & 100\% \\
    \midrule
    % \textbf{BERT-based Models}\\
    % results without entity markers
    % BERT  & 26.22  & 53.96  & 64.00  & 44.72  & 76.88  & 87.72  & 58.94  & 81.50  & 90.40  \\
    % MTB   & 26.18  & 53.38  & 63.74  & 46.98  & 77.34  & 87.44  & 59.52  & 82.40  & 90.14  \\
    % CP    & 40.70  & 55.54  & 64.12  & 47.64  & 78.40  & 87.60  & 73.46  & 86.16  & 91.14  \\
    % $\text{\ourmodel}_{\texttt{BERT}}$ - $L_{EP}$ & 37.14  & 55.62  & 65.06  & 46.96  & 79.06  & 87.70  & 71.88  & 85.04  & 90.70  \\
    % $\text{\ourmodel}_{\texttt{BERT}}$ - $L_{RP}$ & 34.68  & 55.14  & 64.06  & 48.22  & 79.24  & 87.62  & 66.58  & 83.82  & 90.62  \\
    % $\text{\ourmodel}_{\texttt{BERT}}$ & 36.96  & 56.18  & 64.28  & 47.84  & 78.32  & 87.78  & 71.76  & 85.00  & 90.76  \\
    % results with entity markers
    % BERT  & 35.98 & 58.50  & 68.06 & 43.60  & 79.30  & 88.12 & 60.80  & 84.94 & 91.26 \\
    % BERT  & 36.0 & 58.5  & 68.1 & 43.6  & 79.3  & 88.1 & 60.8  & 85.0 & 91.3 \\
    BERT  & 36.0 & 58.5  & 68.1 & 43.6  & 79.3  & 88.1  \\
    % MTB   & 35.72 & 58.76 & 68.20  & 44.16 & 79.24 & 88.24 & 61.82 & 85.86 & 91.46 \\
    % MTB   & 35.7 & 58.8 & 68.2  & 44.2 & 79.2 & 88.2 & 61.8 & 85.9 & 91.5 \\
    MTB   & 35.7 & 58.8 & 68.2  & 44.2 & 79.2 & 88.2  \\
    % CP    & 37.10  & 60.56 & 68.12 & 40.28 & 79.96 & 88.46 & 66.26 & 89.00    & 92.36 \\
    % CP    & \textbf{37.1}  & \textbf{60.6} & 68.1 & 40.3 & 80.0 & \textbf{88.5} & 66.3 & \textbf{89.0}    & \textbf{92.4} \\
    CP    & \textbf{37.1}  & \textbf{60.6} & 68.1 & 40.3 & 80.0 & \textbf{88.5}  \\
    % $\text{\ourmodel}_{\texttt{BERT}}$& 36.52 & 59.66 & 68.48 & 47.88 & 80.06 & 88.04 & 72.00    & 86.68 & 91.62 \\
    % $\text{\ourmodel}_{\texttt{BERT}}$& 36.5 & 59.7 & \textbf{68.5} & \textbf{47.9} & \textbf{80.1} & 88.0 & \textbf{72.0}    & 86.7 & 91.6 \\
    $\text{\ourmodel}_{\texttt{BERT}}$& 36.5 & 59.7 & \textbf{68.5} & \textbf{47.9} & \textbf{80.1} & 88.0  \\
    \midrule
    % \textbf{RoBERTa-based Models}\\
    % RoBERTa & 26.28  & 61.16  & 69.66  & 45.96  & 80.26  & 88.76  & 60.78  & 85.78  & 91.32 \\
    % RoBERTa & 26.3  & 61.2  & 69.7  & 46.0  & 80.3  & 88.8  & 60.8  & 85.8  & 91.3 \\
    RoBERTa & 26.3  & 61.2  & 69.7  & 46.0  & 80.3  & 88.8  \\
    % $\text{\ourmodel}_{\texttt{RoBERTa}}$ & 40.04  & 61.94  & 69.76  & 46.32  & 80.44  & 89.20  & 67.16  & 87.26  & 91.74 \\
    % $\text{\ourmodel}_{\texttt{RoBERTa}}$ & \textbf{40.0}  & \textbf{61.9}  & \textbf{69.8}  & \textbf{46.3}  & \textbf{80.4}  & \textbf{89.2}  & \textbf{67.2}  & \textbf{87.3}  & \textbf{91.7} \\
    $\text{\ourmodel}_{\texttt{RoBERTa}}$ & \textbf{40.0}  & \textbf{61.9}  & \textbf{69.8}  & \textbf{46.3}  & \textbf{80.4}  & \textbf{89.2}  \\
    \bottomrule
    \end{tabular}%
  \caption{Results (test F1) on sentence-level RE (TACRED and SemEval-2010 Task8) on three splits (1\%, 10\% and 100\%).}
  \label{tab:sentre}%
\end{table}%

\paragraph{Document-level RE} For document-level RE, we choose DocRED~\citep{yao2019docred}, which requires reading multiple sentences in a document and synthesizing all the information to identify the relation between two entities. We encode all entities in the same way as in pre-training phase. The relation representations are obtained by adding a bilinear layer on top of two entity representations. We choose the following baselines: (1) \textbf{CNN}~\citep{zeng2014relation}, \textbf{BILSTM} \citep{hochreiter1997long}, \textbf{BERT}~\citep{devlin2018bert} and \textbf{RoBERTa}~\citep{liu2019roberta}, which are widely used as text encoders for relation extraction tasks; (2) \textbf{HINBERT}~\citep{tang2020hin} which employs a hierarchical inference network to leverage the abundant information from different sources; (3) \textbf{CorefBERT}~\citep{ye2020coreferential} which proposes a pre-training method to help BERT capture the coreferential relations in context; (4) \textbf{SpanBERT}~\citep{joshi2019spanbert} which masks and predicts contiguous random spans instead of random tokens; (5) \textbf{ERNIE}~\citep{zhang2019ernie} which incorporates KG information into BERT to enhance entity representations; (6) \textbf{MTB}~\citep{soares2019matching} and \textbf{CP}~\citep{peng2020learning} which introduce sentence-level relation contrastive learning for BERT via distant supervision. For fair comparison, we pre-train these baselines on our constructed pre-training data\footnote{In practice, documents are split into sentences and we only keep within-sentence entity pairs.} based on the implementation released by \citet{peng2020learning}\footnote{\url{https://github.com/thunlp/RE-Context-or-Names}}. From the results shown in Table~\ref{tab:docred}, we can see that: (1) \ourmodel outperforms all baselines significantly on each supervised data size, which demonstrates that \ourmodel could better understand the relations among entities in the document via implicitly considering their complex reasoning patterns in the pre-training; (2) both \textbf{MTB} and \textbf{CP} achieve worse results than \textbf{BERT}, which means sentence-level pre-training, lacking consideration for complex reasoning patterns, hurts PLM's performance on document-level RE tasks to some extent; (3) \ourmodel outperforms baselines by a larger margin on smaller training sets, which means \ourmodel has gained pretty good document-level relation reasoning ability in contrastive learning, and thus obtains improvements more extensively under low-resource settings.
\begin{table}[!t]
  \centering
  \small
    \begin{tabular}{lcc}
    \toprule
    \textbf{Metrics}  & \multicolumn{1}{l}{Macro F1} & \multicolumn{1}{l}{Micro F1} \\
    \midrule
    % \textbf{BERT-based Models}\\
    BERT  & 75.50  & 72.68 \\
    MTB   & 76.37 & 72.94 \\
    CP    & 76.27 & 72.48 \\
    ERNIE & 76.51 & 73.39 \\
    %\midrule
    %$\text{\ourmodel}_{\texttt{BERT}}$ - $L_{EP}$ & 77.27 & 74.04 \\
    %$\text{\ourmodel}_{\texttt{BERT}}$ - $L_{RP}$ & 76.11 & 73.76 \\
    $\text{\ourmodel}_{\texttt{BERT}}$& \textbf{77.85} & \textbf{74.71} \\
    \midrule
    % \textbf{RoBERTa-based Models}\\
    RoBERTa & 79.24 & 76.38 \\
    $\text{\ourmodel}_{\texttt{RoBERTa}}$ & \textbf{80.77} & \textbf{77.04} \\
    \bottomrule
    \end{tabular}%
  \caption{Results on entity typing (FIGER). We report macro F1 and micro F1 on the test set.}
  \label{tab:entity_typing}%
\end{table}%

\paragraph{Sentence-level RE} For sentence-level RE, we choose two widely used datasets: TACRED~\citep{zhang2017position} and SemEval-2010 Task 8~\citep{hendrickx2019semeval}. We insert extra marker tokens to indicate the head and tail entities in each sentence. For baselines, we compare \ourmodel with \textbf{BERT}, \textbf{RoBERTa}, \textbf{MTB} and \textbf{CP}. From the results shown in Table \ref{tab:sentre}, we observe that \ourmodel achieves almost comparable results on sentence-level RE tasks with \textbf{CP}, which means document-level pre-training in \ourmodel does not impair PLMs' performance on sentence-level relation understanding.

\subsection{Entity Typing}

\begin{table}[!t]
  \centering
  \small
    \begin{tabular}{l@{~~~}c@{~~~}c@{~~~}c@{~~~}c@{~~~}c@{~~~}c}
    \toprule
    \textbf{Setting}      & \multicolumn{3}{c}{Standard}    & \multicolumn{3}{c}{Masked} \\
    \midrule
    \textbf{Size}      & 1\%   & 10\%  & 100\% & 1\%   & 10\%  & 100\% \\
    \midrule
    % \multicolumn{3}{l}{\textbf{Traditional Models}}\\
    % FastQA & \multicolumn{2}{c}{-} & 38.00  & \multicolumn{2}{c}{-} & 27.20  \\
    FastQA & \multicolumn{2}{c}{-} & 27.2 & \multicolumn{2}{c}{-} & 38.0 \\
    % BiDAF & \multicolumn{2}{c}{-} & 59.80  & \multicolumn{2}{c}{-} & 49.70  \\
    BiDAF & \multicolumn{2}{c}{-} & 49.7 & \multicolumn{2}{c}{-} & 59.8 \\
    \midrule
    % \multicolumn{3}{l}{\textbf{BERT-based Models}}\\
    % BERT  & 37.88  & 53.09  & 73.13  & 35.79  & 53.74  & 69.53  \\
    BERT  & 35.8  & 53.7  & 69.5 & 37.9  & 53.1  & 73.1 \\
    % CorefBERT & 39.01 & 53.52 & 70.74 & 38.05 & 54.44 & 68.76 \\
    CorefBERT & 38.1 & 54.4 & 68.8 & 39.0 & 53.5 & 70.7 \\
    % SpanBERT & 33.98 & 55.43  & 73.17 & 33.07 & 56.36  & \textbf{70.70} \\
    SpanBERT & 33.1 & 56.4  & \textbf{70.7} & 34.0 & 55.4  & 73.2 \\
    % ERNIE  &       &      &   & \multicolumn{3}{c}{-} \\
    % MTB   & 36.15  & 50.91  & 71.73  & 36.57 & 51.69 & 68.41 \\
    MTB  & 36.6 & 51.7 & 68.4 & 36.2  & 50.9  & 71.7 \\
    % CP    & 34.10  & 47.10  & 69.39  & 34.60  & 50.38 & 67.44 \\
    CP  & 34.6  & 50.4 & 67.4 & 34.1  & 47.1  & 69.4 \\
    %\midrule
    %$\text{\ourmodel}_{\texttt{BERT}}$ - $L_{EP}$ & 37.22  & 52.02  & 68.82  & 72.72  & 37.49 & 52.04 & 65.83 & 68.39 \\
    %$\text{\ourmodel}_{\texttt{BERT}}$ - $L_{RP}$ & \textbf{41.12}  & \textbf{59.84}  & \textbf{71.46}  & \textbf{74.81}  & \textbf{47.47} & \textbf{59.25} & 67.45 & \textbf{70.83} \\
    % $\text{\ourmodel}_{\texttt{BERT}}$& \textbf{40.18}  & \textbf{58.06}  & \textbf{73.87}  & \textbf{46.53}  & \textbf{57.78}  & 69.66  \\
    $\text{\ourmodel}_{\texttt{BERT}}$ & \textbf{46.5}  & \textbf{57.8}  & 69.7 & \textbf{40.2}  & \textbf{58.1}  & \textbf{73.9} \\
    \midrule
    % \multicolumn{3}{l}{\textbf{RoBERTa-based Models}}\\
    % RoBERTa & 41.20  & 58.72  & 75.49  & 37.25  & 57.43  & 70.93  \\
    RoBERTa  & 37.3  & 57.4  & 70.9  & 41.2  & 58.7  & 75.5 \\
    % $\text{\ourmodel}_{\texttt{RoBERTa}}$ & \textbf{46.83}  & \textbf{63.40}  & \textbf{76.56}  & \textbf{47.39}  & \textbf{58.84}  & \textbf{71.21}  \\
    $\text{\ourmodel}_{\texttt{RoBERTa}}$  & \textbf{47.4}  & \textbf{58.8}  & \textbf{71.2}  & \textbf{46.8}  & \textbf{63.4}  & \textbf{76.6} \\
    \bottomrule
    \end{tabular}%
  \caption{Results (accuracy) on the dev set of WikiHop. We test both the standard and masked settings on three splits (1\%, 10\% and 100\%). %We did experiments on three partitions of the original training set (1\%, 10\% and 100\%).
  }
  \label{tab:wikihop}%
\end{table}%

Entity typing aims at classifying entity mentions into pre-defined entity types. We choose FIGER~\citep{ling2015design}, which is a sentence-level entity typing dataset labeled with distant supervision. \textbf{BERT}, \textbf{RoBERTa}, \textbf{MTB}, \textbf{CP} and \textbf{ERNIE} are chosen as baselines. From the results listed in Table \ref{tab:entity_typing}, we observe that, \ourmodel outperforms all baselines, which demonstrates that \ourmodel could better  represent entities and distinguish them in text via both entity-level and relation-level contrastive learning.

\subsection{Question Answering}

\begin{table}[!t]
  \centering
  \small
    \begin{tabular}{l@{~~~}c@{~~~}c@{~~~}c@{~~~}c@{~~~}c@{~~~}c}
    \toprule
    \textbf{Setting} & \multicolumn{2}{c}{SQuAD}     & \multicolumn{2}{c}{TriviaQA}    & \multicolumn{2}{c}{NaturalQA} \\
    \midrule
    \textbf{Size}      & 10\%   & 100\%  & 10\% & 100\%   & 10\%  & 100\% \\
    \midrule
    BERT & 79.7  & \textbf{88.9}  & 60.8  & 70.7  & 68.4  & 78.4 \\
    MTB  & 63.5  & 87.1  & 52.0  & 67.8  & 61.2  & 76.7  \\
    CP  & 69.0  & 87.1   & 52.9  & 68.1  & 63.3  & 77.3  \\
    $\text{\ourmodel}_{\texttt{BERT}}$ & \textbf{81.8}  & 88.9  & \textbf{63.5}  & \textbf{71.9}  & \textbf{70.2}  & \textbf{79.1} \\
    \midrule
    RoBERTa  & 82.9  & \textbf{90.5} & 63.6 & 72.0  & 71.8  & 80.0 \\
    $\text{\ourmodel}_{\texttt{RoBERTa}}$  & \textbf{85.0}  & 90.4  & \textbf{63.6}  & \textbf{72.1}  & \textbf{73.7}  & \textbf{80.5} \\
    \bottomrule
    \end{tabular}%
  \caption{Results (F1) on extractive QA (SQuAD, TriviaQA and NaturalQA) on two splits (10\% and 100\%). Results on 1\% split are left in the appendix.
  }
  \label{tab:extractive_2}%
\end{table}%

Question answering aims to extract a specific answer span in text given a question. We conduct experiments on both multi-choice and extractive QA. We test multiple partitions of the training set.

\paragraph{Multi-choice QA} For Multi-choice QA, we choose WikiHop~\citep{welbl2018constructing}, which requires models to answer specific properties of an entity after reading multiple documents and conducting multi-hop reasoning. It has both standard and masked settings, where the latter setting masks all entities with random IDs to avoid information leakage. We first concatenate the question and documents into a long sequence, then we find all the occurrences of an entity in the documents, encode them into hidden representations and obtain the global entity representation by applying mean pooling on these hidden representations. Finally, we use a classifier on top of the entity representation for prediction.
 % and models are required to answer based only on the context information, without access to entity mention information. 
We choose the following baselines: (1) \textbf{FastQA}~\citep{weissenborn2017making} and \textbf{BiDAF}~\citep{seo2016bidirectional}, which are widely used question answering systems; (2) \textbf{BERT}, \textbf{RoBERTa}, \textbf{CorefBERT}, \textbf{SpanBERT}, \textbf{MTB} and \textbf{CP}, which are introduced in previous sections. From the results listed in Table \ref{tab:wikihop}, we observe that \ourmodel outperforms baselines in both settings, indicating that \ourmodel can better understand entities and their relations in the documents and extract the true answer according to queries. The significant improvements in the masked setting also indicate that \ourmodel can better perform multi-hop reasoning to synthesize and analyze information from contexts, instead of relying on entity mention ``shortcuts''~\citep{jiang2019avoiding}.

% \begin{table}[h]
%   \centering
%   \small
%     \begin{tabular}{l@{~~~}c@{~~~}c@{~~~}c@{~~~}c@{~~~}c@{~~~}c}
%     \toprule
%     \textbf{Setting}      & \multicolumn{3}{c}{TriviaQA}    & \multicolumn{3}{c}{Natural Questions} \\
%     \midrule
%     \textbf{Size}      & 1\%   & 10\%  & 100\% & 1\%   & 10\%  & 100\% \\
%     \midrule
%     BERT  & 28.7  & 60.8  & 70.7 & 31.5  & 68.4  & 78.4 \\
%     MTB & 22.0  & 52.0  & 67.8  & 28.4  & 61.2  & 76.7  \\
%     CP  & 25.6  & 52.9  & 68.1  & 29.4  & 63.3  & 77.3  \\
%     $\text{\ourmodel}_{\texttt{BERT}}$ & \textbf{51.4}  & \textbf{63.5}  & \textbf{71.9} & \textbf{42.9}  & \textbf{70.2}  & \textbf{79.1} \\
%     \midrule
%     RoBERTa  &  &  &  & 34.0  & 71.8  & 80.0 \\
%     $\text{\ourmodel}_{\texttt{RoBERTa}}$  &  &  &  & \textbf{43.0}   & \textbf{73.7}  & \textbf{80.5} \\
%     \bottomrule
%     \end{tabular}%
%   \caption{Results (F1) on Extractive Question Answering (TriviaQA and Natural Questions).
%   }
%   \label{tab:extractive}%
% \end{table}%

\paragraph{Extractive QA} For extractive QA, we adopt three widely-used datasets: SQuAD~\citep{rajpurkar-etal-2016-squad}, TriviaQA~\citep{joshi-etal-2017-triviaqa} and NaturalQA~\citep{kwiatkowski2019natural} in MRQA~\citep{fisch-etal-2019-mrqa} to evaluate \ourmodel in various domains. Since MRQA does not provide the test set for each dataset, we randomly split the original dev set into two halves and obtain the new dev/test set. We follow the QA setting of BERT~\citep{devlin2018bert}: we concatenate the given question and passage into one long sequence, encode the sequence by PLMs and adopt two classifiers to predict the start and end index of the answer.
%given a question $q_1, q_2,...,q_m$ and passages $p_1, p_2, ..., p_n$, we concatenate the question and passages into one long sequence ``\texttt{[CLS]} $q_1, q_2,...,q_m$ \texttt{[SEP]}, $p_1, p_2, ..., p_n$ \texttt{[SEP]}'', which is then encoded by PLMs. Two classifiers are adopted to predict the start and end index of the answer. 
We choose \textbf{BERT}, \textbf{RoBERTa}, \textbf{MTB} and \textbf{CP} as baselines. From the results listed in Table \ref{tab:extractive_2}, we observe that \ourmodel outperforms all baselines, indicating that through the enhancement of entity and relation understanding, \ourmodel is more capable of capturing in-text relational facts and synthesizing information of entities. This ability further improves PLMs for question answering.

\section{Analysis}
% \begin{table}[t!]
%   \centering
%   \small
%     \begin{tabular}{lccc}
%     \toprule
%     \textbf{Dataset}  & \multicolumn{1}{c}{DocRED} & \multicolumn{1}{c}{FIGER}  &
%     \multicolumn{1}{c}{WikiHop} \\
%     \midrule
%     BERT & 54.5  & 72.7  & 73.1 \\
%     \quad-NSP  & 54.6  & 72.6  & 73.3  \\
%     \quad-NSP+$\mathcal{L}_{ED}$ & 55.8  & 73.8 & \textbf{74.8}   \\
%     \quad-NSP+$\mathcal{L}_{RD}^{cross}$ & 54.7  & 72.6  & 72.8  \\
%     \quad-NSP+$\mathcal{L}_{RD}^{single}$ & 55.5  & 73.5  & 72.5  \\
%     \quad-NSP+$\mathcal{L}_{RD}^{both}$ & 55.6  & 74.0  & 72.7  \\
%     $\text{\ourmodel}_{\texttt{BERT}}$ & \textbf{55.9} & \textbf{74.7}  & 73.9 \\
%     \bottomrule
%     \end{tabular}
%   \caption{Ablation study. We report test IgF1 on DocRED, test micro F1 on FIGER and dev accuracy on the masked setting of WikiHop.}
%   \label{tab:ablation}%
% \end{table}%

In this section, we first conduct a suite of ablation studies to explore how $\mathcal{L}_\text{ED}$ and $\mathcal{L}_\text{RD}$ contribute to \ourmodel. Then we give a thorough analysis on how pre-training data's domain / size and methods for entity encoding impact the performance. Lastly, we visualize the entity and relation embeddings learned by \ourmodel.

\subsection{Ablation Study}
To demonstrate that the superior performance of \ourmodel is not owing to its longer pretraining (2500 steps) on masked language modeling, we include a baseline by optimizing $\mathcal{L}_{\text{MLM}}$ only (removing the Next Sentence Prediction (-NSP) loss~\citep{devlin2018bert}). In addition, to explore how $\mathcal{L}_\text{ED}$ and $\mathcal{L}_\text{RD}$ impact the performance, we keep only one of these two losses and compare the results. Lastly, to evaluate how intra-sentence and inter-sentence entity pairs contribute to RD task, we compare the performances of only sampling intra-sentence entity pairs ($\mathcal{L}_{\text{RD}}^{\scriptscriptstyle \mathcal{T}^{+}_{s}, \mathcal{T}^{+}_{s}}$) or inter-sentence entity pairs ($\mathcal{L}_{\text{RD}}^{\scriptscriptstyle \mathcal{T}^{+}_{c}, \mathcal{T}^{+}_{c}}$), and sampling both of them ($\mathcal{L}_\text{RD}$) during pre-training. We conduct experiments on DocRED, WikiHop (masked version) and FIGER. For DocRED and WikiHop, we show the results on 10\% splits and the full results are left in the appendix.

\begin{table}[t]
  \centering
  \small
    \begin{tabular}{lccc}
    \toprule
    \textbf{Dataset}  & \multicolumn{1}{c}{DocRED} & \multicolumn{1}{c}{FIGER}  &
    \multicolumn{1}{c}{WikiHop} \\
    \midrule
    BERT & 44.9  & 72.7  & 53.1 \\
    \quad-NSP  & 45.2  & 72.6  & 53.6  \\
    \quad-NSP+$\mathcal{L}_\text{ED}$ & 47.6  & 73.8 & \textbf{59.8}   \\
    \quad-NSP+$\mathcal{L}_{\text{RD}}^{\scaleto{\mathcal{T}}{3pt}^{\scaleto{+}{3pt}}_{c}, \scaleto{\mathcal{T}}{3pt}^{\scaleto{+}{3pt}}_{c}}$ & 46.4  & 72.6  & 52.2  \\
    \quad-NSP+$\mathcal{L}_{\text{RD}}^{\scaleto{\mathcal{T}}{3pt}^{\scaleto{+}{3pt}}_{s}, \scaleto{\mathcal{T}}{3pt}^{\scaleto{+}{3pt}}_{s}}$ & 47.3  & 73.5  & 51.2  \\
    \quad-NSP+$\mathcal{L}_\text{RD}$ & 48.0  & 74.0  & 52.0  \\
    $\text{\ourmodel}_{\texttt{BERT}}$ & \textbf{48.3} & \textbf{74.7}  & 58.1 \\
    \bottomrule
    \end{tabular}
  \caption{Ablation study. We report test IgF1 on DocRED (10\%), test micro F1 on FIGER and dev accuracy on the masked setting of WikiHop (10\%).}
  \label{tab:ablation}%
\end{table}%

From the results shown in Table \ref{tab:ablation}, we can see that: (1) extra pretraining (-NSP) only contributes a little to the overall improvement. (2) For DocRED and FIGER, either $\mathcal{L}_\text{ED}$ or $\mathcal{L}_\text{RD}$ is beneficial, and combining them further improves the performance; For WikiHop, $\mathcal{L}_\text{ED}$ dominates the improvement while $\mathcal{L}_\text{RD}$ hurts the performance slightly, this is possibly because question answering more resembles the tail entity discrimination process, while the relation discrimination process may have conflicts with it. (3) For $\mathcal{L}_\text{RD}$, both intra-sentence and inter-sentence entity pairs contribute, which demonstrates that incorporating both of them is necessary for PLMs to understand relations between entities in text comprehensively. We also found empirically that when these two auxiliary objectives are only added into the fine-tuning stage, the model does not have performance gain. The reason is that the size and diversity of entities and relations in downstream training data are limited. Instead, pre-training with distant supervision on a large corpus provides a solution for increasing the diversity and quantity of training examples.

\subsection{Effects of Domain Shifting}
%\citet{dontstoppretraining2020} propose that tailoring a pre-trained model to the domain of a target task helps improve its performance. We further investigate how domain shifting impacts \ourmodel's performance. Intuitively, if the data distribution in pre-training phase more resembles that in downstream tasks, PLMs should have better performance. Specifically, we explore two domain shifting factors: entity distribution and relation distribution.
We investigate two domain shifting factors: entity distribution and relation distribution, to explore how they impact \ourmodel's performance.

\paragraph{Entity Distribution Shifting} The entities in supervised datasets of DocRED are recognized by human annotators while our pre-training data is processed by spaCy. Hence there may exist an entity distribution gap between pre-training and fine-tuning. To study the impacts of entity distribution shifting, we fine-tune a BERT model on training set of DocRED for NER tagging and re-tag entities in our pre-training dataset. Then we pre-train \ourmodel on the newly-labeled training corpus (denoted as $\text{\ourmodel}_{\texttt{BERT}}^{\texttt{DocRED}}$). From the results shown in Table \ref{tab:entityshift}, we observe that it performs better than the original \ourmodel, indicating that pre-training on a dataset that shares similar entity distributions with downstream tasks is beneficial.
%In fact, in addition to the supervised training data, DocRED also provides a distantly supervised dataset containing around 100000 documents, whose named entities are recognized using a BERT model finetuned on the supervised training data. Approximately it shares the same entity distribution as the DocRED supervised data. 

\begin{table}[!t]
  \centering
  \small
    \begin{tabular}{lcccc}
    \toprule
    \textbf{Size}      & 1\%  &  10\%  & 100\% \\
    \midrule
    BERT & 28.9 & 44.9  & 54.5  \\
    $\text{\ourmodel}_{\texttt{BERT}}$& 36.0  & 48.3  & 55.9  \\
    $\text{\ourmodel}_{\texttt{BERT}}^{\texttt{DocRED}}$ & \textbf{36.3}  & \textbf{48.6} & \textbf{55.9} \\
    \bottomrule
    \end{tabular}%
  \caption{Effects of pre-training data's entity distribution shifting. We report test IgF1 on DocRED.
  }
  \label{tab:entityshift}%
\end{table}

\begin{figure}[!t]
\centering
\includegraphics[width=0.45\textwidth]{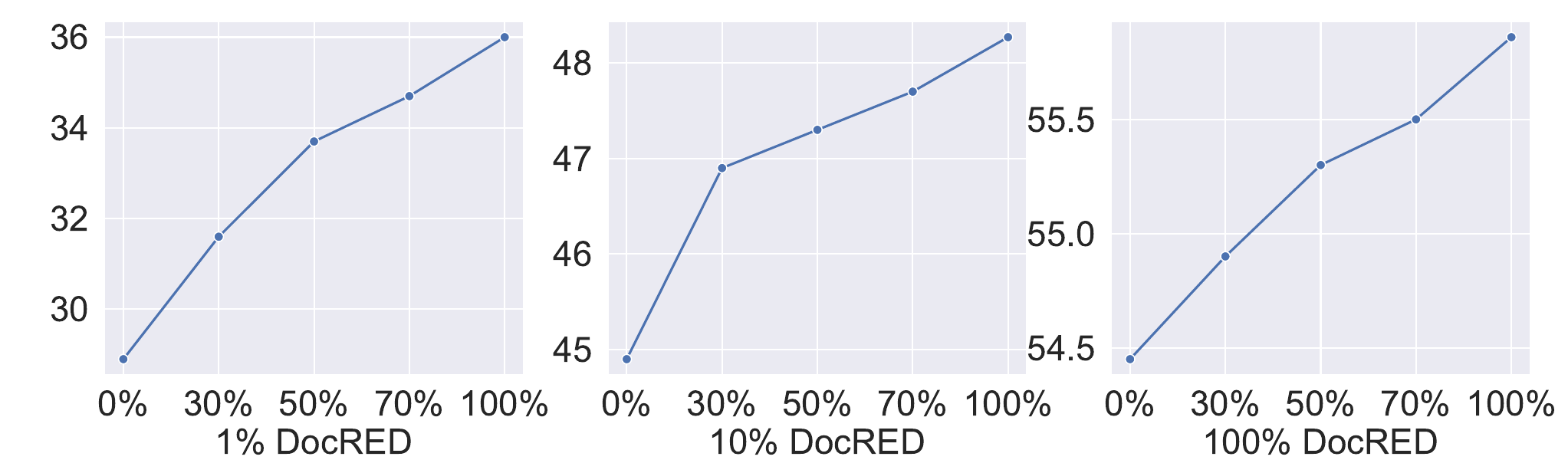}
\caption{Impacts of relation distribution shifting. X axis denotes different ratios of relations, Y axis denotes test IgF1 on different partitions of DocRED.}
\label{fig:rel_analysis}
\end{figure}

\begin{figure}[!t]
\centering
\includegraphics[width=0.45\textwidth]{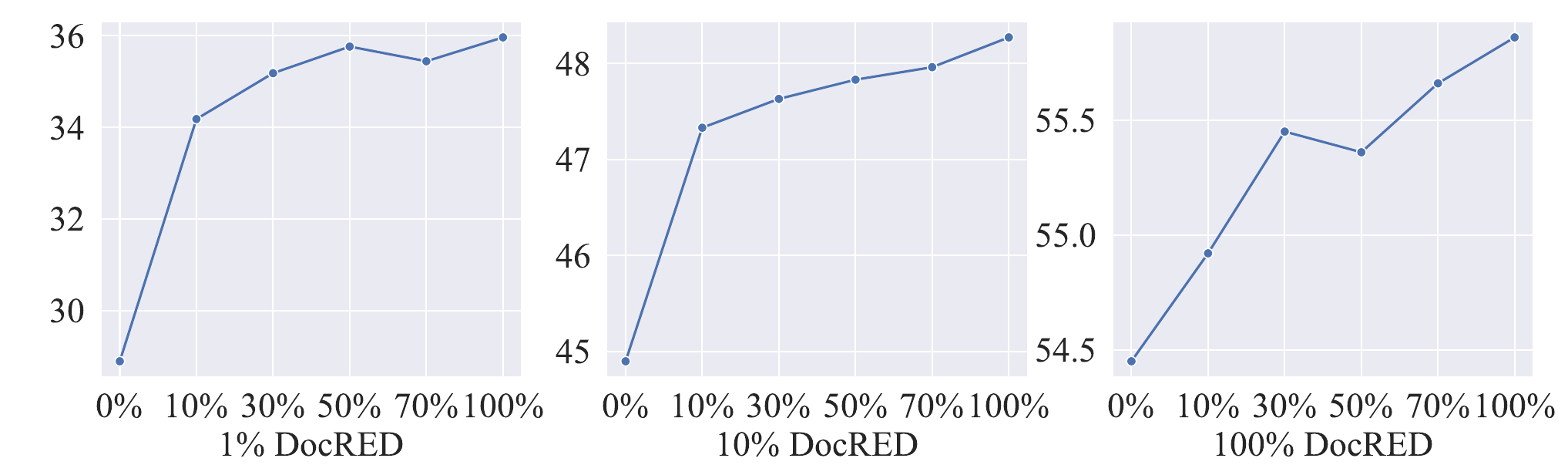}
\caption{Impacts of pre-training data's size. X axis denotes different ratios of pre-training data, Y axis denotes test IgF1 on different partitions of DocRED.}
\label{fig:size_analysis}
\end{figure}

\paragraph{Relation Distribution Shifting} Our pre-training data contains over $4,000$ Wikidata relations. To investigate whether training on a more diverse relation domain benefits \ourmodel, we train it with the pre-training corpus that randomly keeps only $30$\%, $50$\% and $70$\% the original relations, and compare their performances. From the results in Figure \ref{fig:rel_analysis}, we observe that the performance of \ourmodel improves constantly as the diversity of relation domain increases, which reveals the importance of using diverse training data on relation-related tasks. Through detailed analysis, we further find that \ourmodel is less competent at handling unseen relations in the corpus. This may result from the construction of our pre-training dataset: all the relations are annotated distantly through an existing KG with a pre-defined relation set. It would be promising to introduce more diverse relation domains during data preparation in future.

\subsection{Effects of Pre-training Data's Size}

To explore the effects of pre-training data's size, we train \ourmodel on 10\%, 30\%, 50\% and 70\% of the original pre-training dataset, respectively. We report the results in Figure \ref{fig:size_analysis}, from which we observe that with the scale of pre-training data becoming larger, \ourmodel is performing better.

\subsection{Effects of Methods for Entity Encoding}

% \begin{table}[t!]
%   \centering
%   \small
%     \begin{tabular}{lcccc}
%     \toprule
%     \textbf{Size}      & 1\%  &  10\%  & 100\% \\
%     \midrule
%     \textbf{Mean Pool}\\
%     BERT & 28.9 & 44.9  & 54.5  \\
%     $\text{\ourmodel}_{\texttt{BERT}}$& \textbf{36.0}  & \textbf{48.3}  & \textbf{55.9}  \\
%     $\text{\ourmodel}_{\texttt{BERT}}^{\texttt{DocRED}}$ & \textbf{36.3}  & \textbf{48.6} & \textbf{55.9} \\
%     \midrule
%     \textbf{Entity Marker}\\
%     BERT & 23.9  & 44.3  & 55.6  \\
%     $\text{\ourmodel}_{\texttt{BERT}}$& \textbf{34.8}  & \textbf{48.0}  & \textbf{57.6}  \\

%     \bottomrule
%     \end{tabular}%

%   \caption{Results (IgF1) on how entity encoding strategy influences \ourmodel's performance on DocRED. 
%   %+MP means we use mean pooling over all occurrences of each entity to encode it. +SE means we use mean pooling over all start tokens $[S]$ inserted in front of each entity. +DocRED means we pre-train \ourmodel on the distantly supervised dataset provided by DocRED. +x\% denotes the percentage of relation domain we retain.
%   }
%   \label{tab:entityencoding_2}%
% \end{table}%

\begin{table}[t]
  \centering
  \small
    \begin{tabular}{l@{~~~}c@{~~~}c@{~~~}c@{~~~}c@{~~~}c@{~~~}c}
    \toprule
    \textbf{Size}      & \multicolumn{2}{c}{1\%}      & \multicolumn{2}{c}{10\%}      & \multicolumn{2}{c}{100\%} \\
    \midrule
    \textbf{Metrics} & F1    & IgF1  & F1    & IgF1  & F1    & IgF1 \\
    \midrule
    \textbf{Mean Pool}\\
    BERT    & 30.4  & 28.9   & 47.1  & 44.9  & 56.8  & 54.5  \\
    $\text{\ourmodel}_{\texttt{BERT}}$   & 37.8  & 36.0   & 50.8  &  48.3    & 58.2  & 55.9  \\
    $\text{\ourmodel}_{\texttt{BERT}}^{\texttt{DocRED}}$    & \textbf{38.5}  & \textbf{36.3}   & \textbf{51.0}  & \textbf{48.6}   & \textbf{58.2}  & \textbf{55.9}  \\
    \midrule
    \textbf{Entity Marker}\\
    BERT    & 23.0  & 21.8   & 46.5  & 44.3  & 58.0  & 55.6  \\
    $\text{\ourmodel}_{\texttt{BERT}}$   & 34.9  & 33.0   & 50.2  & 48.0   & 59.9  & 57.6  \\
    $\text{\ourmodel}_{\texttt{BERT}}^{\texttt{DocRED}}$   & \textbf{36.9}  & \textbf{34.8}   & \textbf{52.5}  & \textbf{50.3}   & \textbf{60.8} & \textbf{58.4} \\
    \bottomrule
    \end{tabular}%
  \caption{Results (IgF1) on how entity encoding strategy influences \ourmodel's performance on DocRED. We also show the impacts of entity distribution shifting ($\text{\ourmodel}_{\texttt{BERT}}^{\texttt{DocRED}}$ and $\text{\ourmodel}_{\texttt{BERT}}^{\texttt{DocRED}}$) as is mentioned in the main paper.}
  \label{tab:entityencoding}
\end{table}%

For all the experiments mentioned above, we encode each occurrence of an entity by mean pooling over all its tokens in both pre-training and downstream tasks. Ideally, \ourmodel should have consistent improvements on other kinds of methods for entity encoding. To demonstrate this, we try another entity encoding method mentioned by \citet{soares2019matching} on three splits of DocRED (1\%, 10\% and 100\%). Specifically, we insert a special start token \texttt{[S]} in front of an entity and an end token \texttt{[E]} after it. The representation for this entity is calculated by averaging the representations of all its start tokens in the document. To help PLMs discriminate different entities, we randomly assign different marker pairs (\texttt{[S1]}, \texttt{[E1]}; \texttt{[S2]}, \texttt{[E2]}, ...) for each entity in a document in both pre-training and downstream tasks\footnote{In practice, we randomly initialize $100$ entity marker pairs.}. All occurrences of one entity in a document share the same marker pair. We show in Table \ref{tab:entityencoding} that \ourmodel achieves consistent performance improvements for both methods (denoted as \textbf{Mean Pool} and \textbf{Entity Marker}), indicating that \ourmodel is applicable to different methods for entity encoding. Specifically, \textbf{Entity Marker} achieves better performance when the scale of training data is large while \textbf{Mean Pool} is more powerful under low-resource settings. We also notice that training on a dataset that shares similar entity distributions is more helpful for \textbf{Mean Pool}, where $\text{\ourmodel}_{\texttt{BERT}}^{\texttt{DocRED}}$ achieves 60.8 (F1) and 58.4 (IgF1) on 100\% training data.

\subsection{Embedding Visualization}

\begin{figure}[!t]
\centering
\includegraphics[width=0.45\textwidth]{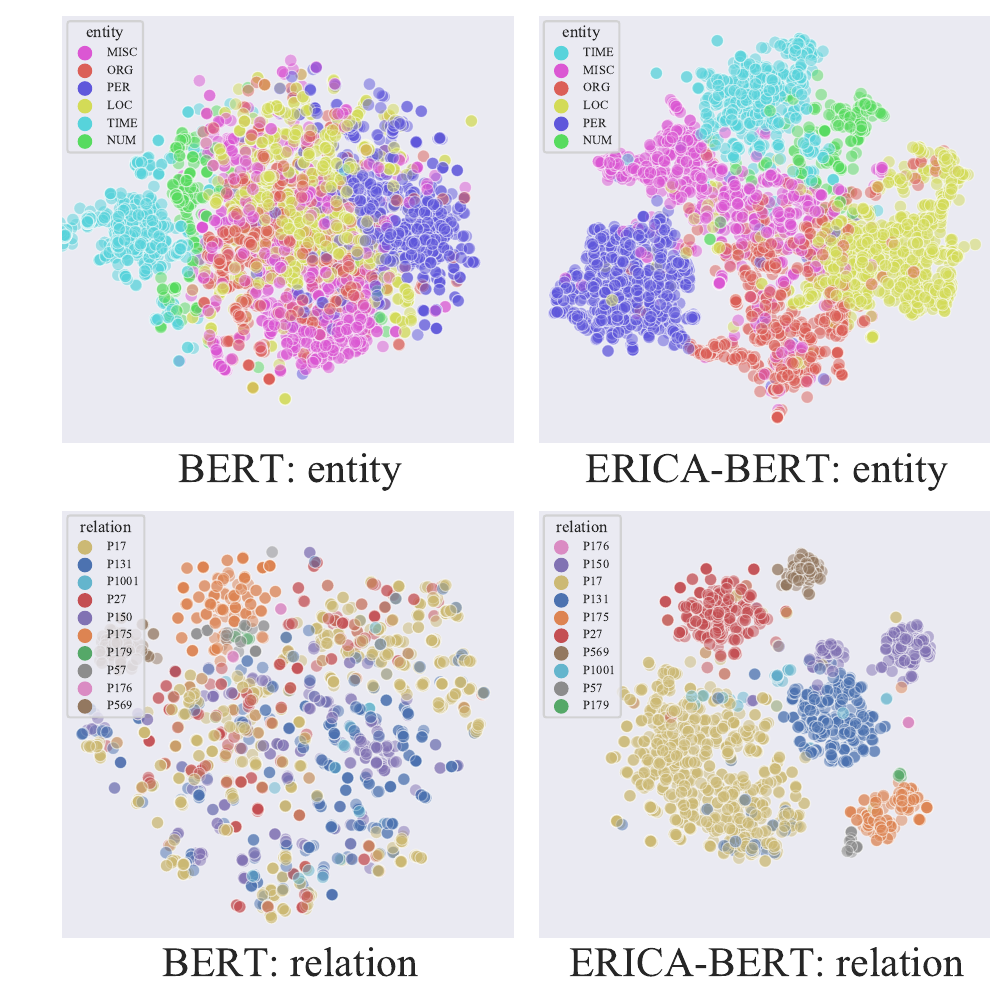}
\caption{t-SNE plots of learned entity and relation embeddings on DocRED comparing BERT and $\text{\ourmodel}_{\texttt{BERT}}$.
}
\label{fig:tsne}
\end{figure}

In Figure \ref{fig:tsne}, we show the learned entity and relation embeddings of BERT and $\text{\ourmodel}_{\texttt{BERT}}$ on DocRED's dev set by t-distributed stochastic neighbor embedding (t-SNE)~\cite{hinton2002stochastic}. We label points with different colors to represent its corresponding category of entities or relations\footnote{(Key, value) pairs for relations defined in Wikidata are: (P176, manufacturer); (P150, contains administrative territorial entity); (P17, country); (P131, located in the administrative territorial entity); (P175, performer); (P27, country of citizenship); (P569, date of birth); (P1001, applies to jurisdiction); (P57, director); (P179, part of the series).} in Wikidata and only visualize the most frequent $10$ relations. From the figure, we can see that jointly training $\mathcal{L}_{\text{MLM}}$ with $\mathcal{L}_\text{ED}$ and $\mathcal{L}_\text{RD}$ leads to a more compact clustering of both entities and relations belonging to the same category. In contrast, only training $\mathcal{L}_{\text{MLM}}$ exhibits random distribution. This verifies that \ourmodel could better understand and represent both entities and relations in the text.
\section{Conclusions}
In this paper, we present \ourmodel, a general framework for PLMs to improve entity and relation understanding via contrastive learning. We demonstrate the effectiveness of our method on several language understanding tasks, including relation extraction, entity typing and question answering. The experimental results show that \ourmodel outperforms all baselines, especially under low-resource settings, which means \ourmodel helps PLMs better capture the in-text relational facts and synthesize information about entities and their relations.

\section*{Acknowledgments}
This work is supported by the National Key Research and Development Program of China (No. 2020AAA0106501) and Beijing Academy of Artificial Intelligence (BAAI). This work is also supported by the Pattern Recognition Center, WeChat AI, Tencent Inc.

\bibliography{tacl2018}
\bibliographystyle{acl_natbib}

\clearpage
\appendix
\section*{Appendices}
\label{apdx}
\section{Training Details for Downstream Tasks}
\begin{table*}[ht]
  \centering
  \small
    \begin{tabular}{lcccccccc}
    \toprule
    \textbf{Dataset}      & MNLI(m/mm)  & QQP   & QNLI  & SST-2 & CoLA  & STS-B & MRPC  & RTE \\
    \midrule
    % \textbf{BERT-based Models}\\
    BERT  & 84.0/84.4 & 88.9  & 90.6  & 92.4  & 57.2  & 89.7  & 89.4  & 70.1  \\
    $\text{\ourmodel}_{\texttt{BERT}}$& 84.5/84.7 & 88.3  & 90.7  & 92.8  & 57.9  & 89.5  & 89.5  & 69.6  \\
    \midrule
    % \textbf{RoBERTa-based Models}\\
    RoBERTa & 87.5/87.3 & 91.9  & 92.8  & 94.8  & 63.6  & 91.2  & 90.2  & 78.7  \\
    $\text{\ourmodel}_{\texttt{RoBERTa}}$ & 87.5/87.5 & 91.6  & 92.6  & 95.0  & 63.5  & 90.7  & 91.5  & 78.5  \\
    \bottomrule
    \end{tabular}%
  \caption{Results on dev sets of GLUE Benchmark. We report matched/mismatched (m/mm) accuracy for MNLI, F1 score for QQP and MRPC, spearman correlation for STS-B and accuracy for other tasks.}
  \label{tab:glue}%
\end{table*}%

In this section, we introduce the training details for downstream tasks (relation extraction, entity typing and question answering). We implement all models based on Huggingface transformers\footnote{\url{https://github.com/huggingface/transformers}}.
\subsection{Relation Extraction}
\paragraph{Document-level Relation Extraction} For document-level relation extraction, we did experiments on DocRED~\citep{yao2019docred}. We modify the official code\footnote{\url{https://github.com/thunlp/DocRED}} for implementation. For experiments on three partitions of the original training set ($1$\%, $10$\% and $100$\%), we adopt batch size of $10, 32, 32$ and training epochs of $400, 400, 200$, respectively. We choose Adam optimizer~\citep{kingma2014adam} as the optimizer and the learning rate is set to $4\times 10^{-5}$. We evaluate on dev set every $20 / 20 / 5$ epochs and then test the best checkpoint on test set on the official evaluation server\footnote{\url{https://competitions.codalab.org/competitions/20717}}.

\paragraph{Sentence-level Relation Extraction} For sentence-level relation extraction, we did experiments on TACRED~\citep{zhang2017position} and SemEval-2010 Task 8~\citep{hendrickx2019semeval} based on the implementation of \citet{peng2020learning}\footnote{\url{https://github.com/thunlp/RE-Context-or-Names}}. We did experiments on three partitions (1\%, 10\% and 100\%) of the original training set. The relation representation for each entity pair is obtained in the same way as in pre-training phase. Other settings are kept the same as \citet{peng2020learning} for fair comparison.

\subsection{Entity Tying}
For entity typing, we choose FIGER~\citep{ling2015design}, whose training set is labeled with distant supervision. We modify the implementation of ERNIE~\citep{zhang2019ernie}\footnote{\url{https://github.com/thunlp/ERNIE}}. In fine-tuning phrase, we encode the entities in the same way as in pre-training phase. We set the learning rate to $3\times 10^{-5}$ and batch size to $256$, and fine-tune the models for three epochs, other hyper-parameters are kept the same as ERNIE.

\subsection{Question Answering}
\paragraph{Multi-choice QA} For multi-choice question answering, we choose WikiHop~\citep{welbl2018constructing}. Since the standard setting of WikiHop does not provide the index for each candidate, we then find them by exactly matching them in the documents. We did experiments on three partitions of the original training data ($1$\%, $10$\% and $100$\%). We set the batch size to $8$ and learning rate to $5\times 10^{-5}$, and train for two epochs.

\begin{table*}[ht]
  \centering
  \small
    \begin{tabular}{lccccccc}
    \toprule
    \textbf{Dataset}  & \multicolumn{3}{c}{DocRED} & \multicolumn{3}{c}{WikiHop (m)}  &
    \multicolumn{1}{l}{FIGER} \\
    \midrule
    \textbf{Size}  & 1\%  & 10\%  & 100\% & 1\%   & 10\%  & 100\% & 100\% \\
    \midrule
    BERT & 28.9  & 44.9  & 54.5  & 37.9  & 53.1  & 73.1  & 72.7  \\
    \quad-NSP  & 30.1  & 45.2  & 54.6  & 38.2 & 53.6   & 73.3  & 72.6  \\
    \quad-NSP+$\mathcal{L}_{\text{ED}}$ & 34.4  & 47.6  & 55.8  & \textbf{41.1} & \textbf{59.8} & \textbf{74.8} &  73.8  \\
    \quad-NSP+$\mathcal{L}_{\text{RD}}^{\scaleto{\mathcal{T}}{3pt}^{\scaleto{+}{3pt}}_{c}, \scaleto{\mathcal{T}}{3pt}^{\scaleto{+}{3pt}}_{c}}$ & 34.8  & 46.4  & 54.7  & 37.4  & 52.2  & 72.8  & 72.6  \\
    \quad-NSP+$\mathcal{L}_{\text{RD}}^{\scaleto{\mathcal{T}}{3pt}^{\scaleto{+}{3pt}}_{s}, \scaleto{\mathcal{T}}{3pt}^{\scaleto{+}{3pt}}_{s}}$ & 33.9  & 47.3  & 55.5  & 38.0  & 51.2  & 72.5  & 73.5  \\
    \quad-NSP+$\mathcal{L}_{\text{RD}}$ & 35.9  & 48.0  & 55.6  & 37.2  & 52.0  & 72.7  & 74.0  \\
    $\text{\ourmodel}_{\texttt{BERT}}$ & \textbf{36.0} & \textbf{48.3} & \textbf{55.9} & 40.2  & 58.1  & 73.9  & \textbf{74.7} \\
    \bottomrule
    \end{tabular}
  \caption{Full results of ablation study. We report test IgF1 on DocRED, dev accuracy on the masked (m) setting of WikiHop and test micro F1 on FIGER.}
  \label{tab:ablation_full}%
\end{table*}%

\paragraph{Extractive QA} For extractive question answering, we adopt MRQA~\citep{fisch-etal-2019-mrqa} as the testbed and choose three datasets: SQuAD~\citep{rajpurkar-etal-2016-squad}, TriviaQA~\citep{joshi-etal-2017-triviaqa} and NaturalQA~\citep{kwiatkowski2019natural}. We adopt Adam as the optimizer, set the learning rate to $3\times 10^{-5}$ and train for two epochs. In the main paper, we report results on two splits (10\% and 100\%) and results on 1\% are listed in Table \ref{tab:extractive_3}.

\begin{table}[ht]
  \centering
  \small
    \begin{tabular}{l@{~~~}c@{~~~}c@{~~~}c}
    \toprule
    \textbf{Setting} & \multicolumn{1}{c}{SQuAD}     & \multicolumn{1}{c}{TriviaQA}    & \multicolumn{1}{c}{NaturalQA} \\
    \midrule
    BERT & 15.8  & 28.7  & 31.5  \\
    MTB  & 11.2  & 22.0  & 28.4  \\
    CP  & 12.5  & 25.6   & 29.4  \\
    $\text{\ourmodel}_{\texttt{BERT}}$ & \textbf{51.3}  & \textbf{51.4}  & \textbf{42.9}   \\
    \midrule
    RoBERTa  & 22.1  & 40.6 & 34.0   \\
    $\text{\ourmodel}_{\texttt{RoBERTa}}$  & \textbf{57.6}  & \textbf{51.3}  & \textbf{57.6}    \\
    \bottomrule
    \end{tabular}%
  \caption{Results (F1) on extractive QA (SQuAD, TriviaQA and NaturalQA) on 1\% split.
  }
  \label{tab:extractive_3}%
\end{table}%

\section{Generalized Language Understanding (GLUE)}
The General Language Understanding Evaluation (GLUE) benchmark~\citep{wang2018glue} provides several natural language understanding tasks, which is often used to evaluate PLMs. To test whether $\mathcal{L}_{\text{ED}}$ and $\mathcal{L}_{\text{RD}}$ impair the PLMs' performance on these tasks, we compare BERT, $\text{\ourmodel}_{\texttt{BERT}}$, RoBERTa and $\text{\ourmodel}_{\texttt{RoBERTa}}$. We follow the widely used setting and use the \texttt{[CLS]} token as representation for the whole sentence or sentence pair for classification or regression. Table~\ref{tab:glue} shows the results on dev sets of GLUE Benchmark. It can be observed that both $\text{\ourmodel}_{\texttt{BERT}}$ and $\text{\ourmodel}_{\texttt{RoBERTa}}$ achieve comparable performance than the original model, which suggests that jointly training $\mathcal{L}_{\text{ED}}$ and $\mathcal{L}_{\text{RD}}$ with $\mathcal{L}_{\text{MLM}}$ does not hurt PLMs' general ability of language understanding. 

% \section{Additional results on sentence-level RE (Wiki80)}

% \begin{table}[htbp]
%   \centering
%   \small
%     \begin{tabular}{lccc}
%     \toprule
%     \textbf{Size}  & 1\%   & 10\%  & 100\% \\
%     \midrule
%     BERT  & 60.8  & 85.0  & 91.3  \\
%     MTB   & 61.8  & 85.9  & 91.5  \\
%     CP    & 66.3  & \textbf{89.0}  & \textbf{92.4}  \\
%     $\text{\ourmodel}_{\texttt{BERT}}$ & \textbf{72.0} & 86.7 & 91.6  \\
%     \midrule
%     RoBERTa & 60.8  & 85.8  & 91.3  \\
%     $\text{\ourmodel}_{\texttt{RoBERTa}}$  & \textbf{67.2}  & \textbf{87.3}  & \textbf{91.7}  \\
%     \bottomrule
%     \end{tabular}%
%   \caption{Results (accuracy) on test set of Wiki80.}
%   \label{tab:wiki80}%
% \end{table}%

% We additionally did experiments on Wiki80~\citep{han2019opennre}, which is a sentence-level relation extraction dataset. We adopt the same experimental setting as described before. From the results in Table \ref{tab:wiki80}.

\section{Full results of ablation study}
Full results of ablation study (DocRED, WikiHop and FIGER) are listed in Table \ref{tab:ablation_full}.

\section{Joint Named Entity Recognition and Relation Extraction}

\begin{table}[t]
  \centering
  \small
    \begin{tabular}{lcccc}
    \toprule
    \multicolumn{1}{l}{\multirow{2}[4]{*}{\textbf{Model}}} & \multicolumn{2}{c}{CoNLL04} & \multicolumn{2}{c}{ADE} \\
\cmidrule{2-5}          & \multicolumn{1}{c}{NER} & \multicolumn{1}{c}{RE} & \multicolumn{1}{c}{NER} & \multicolumn{1}{c}{RE} \\
    \midrule
    BERT  &   88.5    & 70.3      &   89.2    &  79.2 \\
    $\text{\ourmodel}_{\texttt{BERT}}$     &  \textbf{89.3}     &  \textbf{71.5}     &   \textbf{89.5}    &  \textbf{80.2} \\
    \midrule
    RoBERTa &   89.8    &   72.0    &  89.7     &  81.6 \\
    $\text{\ourmodel}_{\texttt{RoBERTa}}$     &  \textbf{90.0}     &  \textbf{72.8}       & \textbf{90.2}      &  \textbf{82.4} \\
    \bottomrule
    \end{tabular}%
  \caption{Results (F1) on joint NER\&RE.}
  \label{tab:jointNERRE}%
\end{table}%

Joint Named Entity Recognition (NER) and Relation Extraction (RE) aims at identifying entities in text and the relations between them. We adopt SpERT~\citep{DBLP:journals/corr/abs-1909-07755} as the base model and conduct experiments on two datasets: CoNLL04~\citep{roth-yih-2004-linear} and ADE~\citep{gurulingappa2012development} by replacing the base encoders (BERT and RoBERTa) with $\text{\ourmodel}_{\texttt{BERT}}$ and $\text{\ourmodel}_{\texttt{RoBERTa}}$, respectively. We modify the implementation of SpERT\footnote{\url{https://github.com/markus-eberts/spert}} and keep all the settings the same. From the results listed in Table \ref{tab:jointNERRE}, we can see that \ourmodel outperforms all baselines, which again demonstrates the superiority of \ourmodel in helping PLMs better understand and represent both entities and relations in text.

\end{document}